\documentclass{article}

\usepackage{microtype}
\usepackage{booktabs}
\usepackage{multirow}
\usepackage{graphicx}
\usepackage{amssymb}
\usepackage{pifont}
\usepackage{subcaption}
\usepackage{booktabs} %

\usepackage{hyperref}

\usepackage[preprint]{icml2026}

\usepackage{amsmath}
\usepackage{amssymb}
\usepackage{mathtools}
\usepackage{amsthm}

\usepackage[capitalize,noabbrev]{cleveref}

\theoremstyle{plain}

\theoremstyle{definition}

\theoremstyle{remark}

\usepackage{xspace}

\usepackage[dvipsnames,svgnames]{xcolor}

\definecolor{darkgreen}{rgb}{0.0, 0.75, 0.0} %

\definecolor{darkyellow}{rgb}{0.75, 0.75, 0.0} %

\usepackage{colortbl}

\makeatletter
\DeclareRobustCommand\onedot{\futurelet\@let@token\@onedot}
\def\@onedot{\ifx\@let@token.\else.\null\fi\xspace}

\def\eg{\emph{e.g}\onedot} 
\def\ie{\emph{i.e}\onedot}

\makeatother

\crefname{section}{Sec.}{Secs.}
\Crefname{section}{Section}{Sections}
\crefname{table}{Tab.}{Tabs.}
\Crefname{table}{Table}{Tables}
\crefname{figure}{Fig.}{Figs.}
\Crefname{figure}{Figure}{Figures}
\crefname{appendix}{App.}{Apps.}
\Crefname{appendix}{Appendix}{Appendices}

\newcommand{\ignore}[1]{}

\newif\ifdrafting
\draftingtrue %
\ifdrafting
    \newcommand{\todotxt}[1]{{\leavevmode\color[rgb]{1,0,0}[TODO: #1]}}
    \newcommand{\ds}[1]{{\leavevmode\color[rgb]{0.8,0,0.8}[Deqing: #1]}}
    \newcommand{\fc}[1]{{\leavevmode\color[rgb]{0.18,0.64,0.37}[Forrester: #1]}}
    \newcommand{\cih}[1]{{\leavevmode\color[rgb]{0,0.5,0}[Charles: #1]}}
    \newcommand{\jh}[1]{{\leavevmode\color[rgb]{0.8, 0.2, 0}[Junhwa: #1]}}
    \newcommand{\jz}[1]{{\leavevmode\color[rgb]{0,0.6,0.8}[Junyi: #1]}}
    \newcommand{\mh}[1]{{\leavevmode\color[rgb]{0,0.8,0.8}[MHY: #1]}}
    \newcommand{\chen}[1]{{\leavevmode\color[rgb]{0,0.8,0}[Chen: #1]}}
    
\else
    \newcommand{\todotxt}[1]{}
    \newcommand{\ds}[1]{}
    \newcommand{\fc}[1]{}
    \newcommand{\cih}[1]{}
    \newcommand{\jh}[1]{}
    \newcommand{\jz}[1]{}
    \newcommand{\mh}[1]{}
    \newcommand{\chen}[1]{}
    
\fi

\usepackage{enumitem}
\setlist[itemize]{noitemsep, topsep=0pt}
\setlist[enumerate]{noitemsep, topsep=0pt}

\newcommand{\pithree}{$\pi^3$}
\newcommand{\ourmethod}{LoGeR}

\usepackage[textsize=tiny]{todonotes}
\definecolor{myurlblue}{rgb}{0.0, 0.3, 0.8} 
\hypersetup{urlcolor=myurlblue}

\icmltitlerunning{\ourmethod{}: Long-Context Geometric Reconstruction with Hybrid Memory}
\makeatletter

\renewcommand{\printAffiliationsAndNotice}[1]{%
    \global\icml@noticeprintedtrue
}
\makeatother
\begin{document}

\twocolumn[{
\vspace*{-6mm}
  \icmltitle{\LARGE \ourmethod{}: Long-Context Geometric Reconstruction\\ with Hybrid Memory}

  \icmlsetsymbol{equal}{*}

\begin{icmlauthorlist}
\vspace*{-2mm}
\fontsize{10.5pt}{12.5pt}\selectfont
        \textbf{Junyi Zhang}$^{1,2}$\quad
        \textbf{Charles Herrmann}$^{1,*}$\quad
        \textbf{Junhwa Hur}$^{1,*}$\quad
        \textbf{Chen Sun}$^{1}$\quad
        \textbf{Ming-Hsuan Yang}$^{1}$ \\
        \textbf{Forrester Cole}$^{1}$\quad
        \textbf{Trevor Darrell}$^{2}$\quad
        \textbf{Deqing Sun}$^{1,\dagger}$ \\
        \vspace{0.5em} 
        $^{1}$Google DeepMind \quad $^{2}$UC Berkeley
\end{icmlauthorlist}
    
    \printAffiliationsAndNotice{}

  \vskip 4.5mm

    \begin{center}
    \centering
    \includegraphics[width=\linewidth]{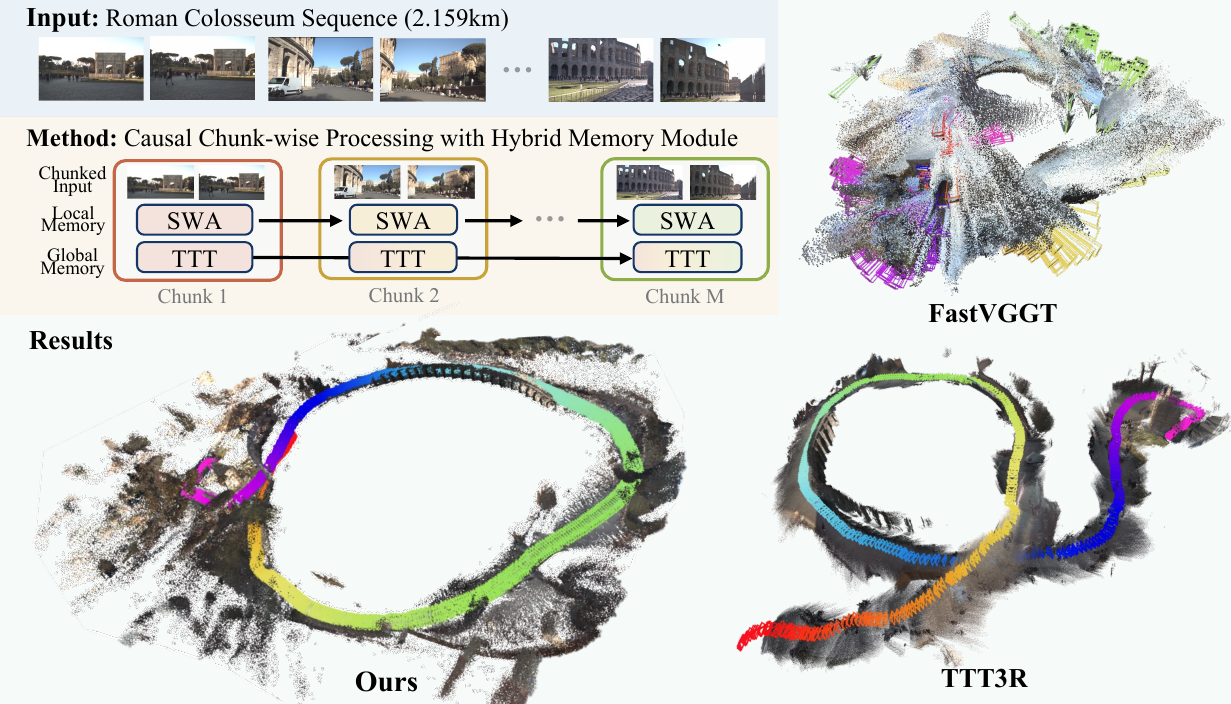}
    \vspace{-1.5em}
    \captionof{figure}{\textbf{Our proposed method and visual comparison.} For very long videos, we advocate chunk-based processing where bidirectional attention handles intra-chunk reasoning with inter-chunk alignment handled by our proposed \textit{hybrid memory module}, composing Sliding Window Attention (SWA) for detailed local memory and Test-Time Training (TTT) for compressed global context. \ourmethod{} shows clear improvement over prior methods on expansive VBR~\cite{brizi2024vbr} dataset, yielding superior loop closures and finer geometric details.
}
    \label{fig:teaser}
    \vspace{3mm}
  \end{center}
}
]

\printAffiliationsAndNotice{}  %

\let\thefootnote\relax\footnotetext{\fontsize{8pt}{10pt}\selectfont $^*$Project leads,  $^\dagger$Direction lead}

\begin{abstract}

\vspace{-0.5em}

Feedforward geometric foundation models achieve strong short-window reconstruction, yet scaling them to minutes-long videos is bottlenecked by quadratic attention complexity or limited effective memory in recurrent designs. We present \textbf{\ourmethod{}} (\textbf{Lo}ng-context \textbf{Ge}ometric \textbf{R}econstruction), a novel architecture that scales dense 3D reconstruction to extremely long sequences without post-optimization. \ourmethod{} processes video streams in chunks, leveraging strong bidirectional priors for high-fidelity intra-chunk reasoning. To manage the critical challenge of coherence across chunk boundaries, we propose a learning-based \textit{hybrid memory module}. This dual-component system combines a parametric Test-Time Training (TTT) memory to anchor the global coordinate frame and prevent scale drift, alongside a non-parametric Sliding Window Attention (SWA) mechanism to preserve uncompressed context for high-precision adjacent alignment. Remarkably, this memory architecture enables \ourmethod{} to be trained on sequences of 128 frames, and generalize up to thousands of frames during inference. Evaluated across standard benchmarks and a repurposed VBR dataset with sequences of up to 19k frames, \ourmethod{} substantially outperforms prior state-of-the-art feedforward methods---reducing ATE on KITTI by over 74\%---and achieves robust, globally consistent reconstruction over unprecedented horizons. Webpage: \url{https://LoGeR-project.github.io/}. 

\end{abstract}

\section{Introduction}
\label{sec:intro}

\vspace{-0.375em}

Large-scale dense 3D reconstruction has long been a central goal in computer vision, underpinning many important applications, from holistic scene understanding to generative world-building. For decades, classical optimization frameworks dominated this landscape; and while capable of reconstructing entire cities, they rely on computationally intensive offline processes and often falter on sparse or texture-less inputs. Recently, geometric foundation models —such as DUSt3R~\cite{wang2024dust3r}, 
MonST3R~\cite{zhang2024monst3r}, VGGT~\cite{wang2025vggt}, and $\pi^3$~\cite{wang2025pi}— are driving a paradigm shift. By distilling complex geometric priors from vast datasets, these models enable robust, feedforward inference even where classical methods fail. However, a critical gap remains. While classical pipelines scale to cities, feedforward models are currently confined to bounded scenes.

The primary obstacle to this scaling is twofold: a fundamental ``context wall'' inherent in current architectures and a severe ``data wall'' during training. Architecturally, while bidirectional attention is essential for learning complex geometric priors, its quadratic complexity restricts its use to short-context windows. From a data perspective, current models are predominantly trained on short-context ``bubbles'' (dozens to over a hundred frames), leaving them fundamentally ill-equipped to integrate long-range dependencies at inference time (thousands to tens of thousands of frames). Consequently, inference-time heuristics like FastVGGT~\cite{shen2025fastvggt}, which successfully mitigate memory bottlenecks, still fail to generalize to the large-scale VBR dataset~\cite{brizi2024vbr}, as shown in~\cref{fig:teaser,fig:data_scenescale}. 

To overcome this data wall without waiting for the curation of massive long-horizon datasets with the requisite diversity and accuracy, we argue that end-to-end chunk-wise processing is a practical and effective strategy. Decomposing the sequence ensures that local inferences remain ``in-distribution'' relative to existing short-context training data and allows us to leverage strong bidirectional, attention-based baselines such as \pithree{}~\cite{wang2025pi}. However, this paradigm introduces a critical new challenge: managing memory and coherence across chunk boundaries. Achieving high-fidelity dense 3D reconstruction over long sequences requires balancing three distinct levels of coherence: intra-window details, uncompressed context for high-precision local alignment, and global structural integrity over long ranges. Existing sequential models fail to balance these conflicting needs. For instance, recurrent approaches like CUT3R~\cite{wang2025continuous} compress all temporal context into a single lossy hidden state, sacrificing the high-precision dense information needed for seamless adjacent alignment. Conversely, naive deterministic stitching preserves local detail but lacks the long-range memory required to prevent scale drift. 

These failures suggest that a single memory strategy is fundamentally insufficient. To bridge this gap, we propose a learning-based \textit{hybrid memory module} that maintains multi-scale geometric coherence within a fixed compute budget. As shown in~\cref{fig:teaser}, we formalize this via a dual-component memory system: a \textit{non-parametric} sliding window attention (SWA) mechanism that preserves uncompressed local features and high-fidelity detail for the most recent chunks, and a \textit{parametric} associative memory (Test-Time Training~\cite{sun2024learning}) that compresses global context. This hybrid design effectively decouples these tasks: the long-range parametric memory anchors the global coordinate frame to prevent scale drift, while the short-range non-parametric memory ensures seamless, high-precision transitions. 

Furthermore, to evaluate these mechanisms at an unprecedented scale, we introduce a long-context reconstruction benchmark derived from the VBR dataset~\cite{brizi2024vbr}, spanning up to 19k frames and 11.5~km in trajectory. Remarkably, our architectural design enables the model to be trained on sequences of {128 frames}, generalize seamlessly up to {1k frames} during inference, and scale effectively up to {19k frames} with periodic state resets and an optional feedforward pose-alignment. As shown in~\cref{fig:teaser}, \ourmethod{} effectively breaks both the context and data walls. It substantially outperforms prior state-of-the-art feedforward methods on KITTI, reducing the Absolute Trajectory Error (ATE) from \textbf{72.86 to 18.65}, and achieves a \textbf{55.2\%} relative improvement on the extremely long sequences of our VBR benchmark.

\begin{table}[t]
\centering
\caption{\textbf{Architectural trade-offs in sequence modeling.} Our hybrid memory module achieves a balance, preserving lossless local geometric details while maintaining global structural consistency at a linear computational cost with respect to sequence length.}
\label{tab:architecture_comparison}
\vspace{-0.25em}
\resizebox{0.95\columnwidth}{!}{%
\begin{tabular}{l c c c}
\toprule
\textbf{Mechanism} & \textbf{Compute Cost} & \textbf{Local Context} & \textbf{Global Context} \\
\midrule
Full Attention & $\mathcal{O}(N^2)$ & Lossless & Lossless \\
Sliding Window Attn. & $\mathcal{O}(N)$ & Lossless & Limited \\
TTT / Linear Attn. & $\mathcal{O}(N)$ & Compressed & Compressed \\
\midrule
\rowcolor{gray!10} \textbf{Ours (Hybrid Memory)} & $\mathcal{O}(N)$ & \textbf{Lossless} & \textbf{Compressed} \\
\bottomrule
\end{tabular}%
}
\vspace{-1.5em}
\end{table}

\vspace{-0.25em}

\vspace{-0.25em}
\section{Related Work}

\vspace{-0.25em}

\noindent \textbf{Learning-based visual SLAM.}
Recent learning-based visual SLAM methods demonstrate outperforming classical approaches~\citep{mur2015orb,mur2017orb,engel2014lsd,newcombe2011dtam} by learning strong 3D priors from training datasets~\citep{teed2021droid,lipson2024deep,li2025megasam} or leveraging powerful pretrained visual geometry models~\citep{deng2025vggt,maggio2025vggt}.
Yet these methods retain (relatively) expensive backends for graph construction, loop closure, and global optimization.
While slower than feedforward methods, such SLAM-based pipelines represent the only viable approach for maintaining coherence over long contexts. 
Here, we provide a fully feedforward alternative that outperforms strong SLAM systems in our long-context evaluation, without relying on any backend optimization.

\noindent \textbf{Feedforward 3D reconstruction.}
Recent work employs feedforward models for 3D reconstruction, directly outputting pointmaps in a canonical space~\citep{wang2024dust3r, leroy2024grounding,zhang2024monst3r,wang2025vggt, wang2025pi}.
Despite achieving high-quality results, these methods can only process a limited number of input frames due to high memory cost from global attention layers, hindering real-world applications (\eg robotics or autonomous driving) that require global consistency over long streams.

\noindent \textbf{Long sequence reconstruction.}
Recent feedforward approaches tackle long sequence reconstruction using external spatial memory~\citep{chen2025long3r,wang20243d,wu2025point3r,lan2025stream3r}, RNN-style persistent state~\citep{wang2025continuous}, or efficiency-oriented VGGT variants with sparse~\citep{shen2025fastvggt,yuan2026infinitevggt} or causal~\citep{zhuo2025streaming,cheng2026longstream} attention.
However, their lack of demonstrated results on exceptionally long sequences (\eg, minute-long) leaves their scalability in sequence lengths unverified.
Concurrently, TTT3R~\citep{chen2025ttt3r} introduces a confidence-based update for single-frame streaming. However, this frame-wise linear approach lacks the expressivity to capture complex temporal context or leverage the powerful multi-frame reasoning of bidirectional backbones~\citep{wang2025vggt,wang2025pi}.

To overcome this, \ourmethod{} processes streams in a chunk-wise manner. Unlike frame-wise updates, our design uniquely preserves the high-fidelity multi-frame reasoning power of bidirectional backbones while leveraging a hybrid memory mechanism to propagate information across chunks.

\noindent \textbf{Memory for long-context modeling.} 
Efficient long-sequence architectures revisit recurrent/state-space models~\citep{gu2021efficiently,gu2024mamba} and linear attention~\citep{katharopoulos2020transformers,schlag2021linear} to reduce the quadratic cost of Transformers. A complementary line of work uses local attention (\eg, sliding-window attention)~\citep{beltagy2020longformer,zaheer2020big} to preserve high-fidelity short-range interactions, but such designs have limited access to global context. Recently, fast-weight mechanisms such as Test-Time Training (TTT)~\citep{sun2024learning,zhang2025test} treat memory as an evolving parameter state, enabling compact accumulation of long-range information, albeit with an inherent compression trade-off. As summarized in~\cref{tab:architecture_comparison}, these individual mechanisms inherently trade off computational cost, local context preservation, and global context accessibility.

While hybrid architectures in language models frequently combine these efficient mechanisms with full global attention to maintain performance (\eg, Longformer~\citep{beltagy2020longformer} or Jamba~\citep{lenz2025jamba}), the extreme token density in dense vision prediction tasks makes this computationally prohibitive. Our method therefore introduces a hybrid memory that remains linear in sequence length, synergizing non-parametric SWA for precise adjacent alignment with parametric TTT for long-range global consistency.

\noindent \textbf{Dataset.} 
Existing training datasets include both real (ARKitScenes, DL3DV~\citep{ling2024dl3dv}, MegaDepth~\cite{li2018megadepth}, ScanNet~\cite{dai2017scannet}, Waymo~\cite{schwall2020waymo}) and synthetic ones (HyperSim~\cite{roberts2021hypersim}, Spring~\cite{mehl2023spring}, TartanAir~\citep{wang2020tartanair}, TartanAirV2~\citep{patel2025tartanground}, UnReal4K~\cite{aleotti2021neural}, Virtual KITTI 2~\citep{cabon2020vkitti2}, OmniWorld-Game~\cite{zhou2025omniworld}). ScanNet and TUM are popular evaluation benchmarks. 
However, these datasets are fundamentally limited in scale—both temporally and spatially. Even though some sequences may contain up to a thousand frames, they typically capture spatially bounded, room-scale environments. To truly evaluate long-context geometric reconstruction in expansive spaces, we adapt the VBR dataset for our evaluation. Spanning from 8,815 to 18,846 frames and covering extensive spatial trajectories up to 11.5~km, this benchmark presents a significant challenge for existing methods.
Our chunk-wise formulation allows models trained on these limited datasets to generalize effectively to the expansive VBR benchmark.

\section{Preliminary}

We first provide the technical background for our approach.

\noindent \textbf{3D reconstruction models.} Existing methods often employ bidirectional transformer architectures to capture complex geometric information. Given a set of $N$ multi-view images or video frames as input $\mathbf{I} = \{I_i\}_{i=1}^N$, where $I_i \in \mathbb{R}^{H \times W \times 3}$,
the transformer predicts geometry as local pointmaps $\mathbf{P}_i \in \mathbb{R}^{H \times W \times 3}$ in \textit{local camera coordinates}, alongside camera poses $\mathbf{c}_i \in \mathbb{R}^{4 \times 4}$ in \textit{global world coordinates}. {While effective for short context, the transformer-based methods are limited to bounded scenes.}

\noindent \textbf{Test-Time Training (TTT).} TTT~\cite{sun2024learning} introduces an architectural component designed to store useful context from prior forward calls of the model.
It achieves this using \emph{fast weights}, a set of parameters updated during both train and inference time. This contrasts with the model's base parameters, or \emph{slow weights}, which remain frozen during inference.

Let $\mathbf{x} = [x_1, x_2, \dots, x_N]$ be a 1D sequence of tokens where each token $x_i \in \mathbb{R}^d$.
Following standard attention mechanism, each input token $x_i$ is
then projected into queries $q_i$, keys $k_i$, and values $v_i$, with $q_i, k_i, v_i \in \mathbb{R}^d$.
Formally, TTT defines a neural network $f_W(\cdot): \mathbb{R}^d \rightarrow \mathbb{R}^d$ parameterized by the fast weights $W$, which are then utilized in two operations.
\vspace{-0.4em}
\begin{equation}
\textbf{Update operation:} \quad W \leftarrow W - \eta \nabla_{W} \mathcal{L}\big(f_W(\mathbf{k}), \mathbf{v}\big)
\label{eq:ttt_update}
\end{equation}
where $\eta$ is the learning rate and $\mathcal{L}(\cdot,\cdot)$ is a loss function between the transformed key $f_W(k)$ and the value $v$. This loss encourages the function $f_W$ to link keys with corresponding values, conceptually trying to encode the KV cache into the $W$ matrix as a form of neural memory.
\begin{equation}
\textbf{Apply operation:}\quad
o = f_W(\mathbf{q}),\quad 
\label{eq:ttt_apply}
\end{equation}
where the updated fast weights $W$ process the query $q$ to compute the output vector $o$.
The per-token TTT layer iteratively performs the update and applies operations on each token $x_i$ in sequence.
Following previous literature, we implement the fast weight architecture using SwiGLU layer and employ the Muon optimizer~\cite{jordan2024muon} for the test-time updates. 
{While TTT can compress and propagate information for long contexts with linear complexity, its compressed memory is inherently lossy. Next, we will introduce our hybrid-memory approach that uses different memory mechanisms to handle geometric information at various time scales over a long context.}

\label{sec:arch}
\begin{figure}[t]
\centering
    \includegraphics[width=1.0\linewidth]{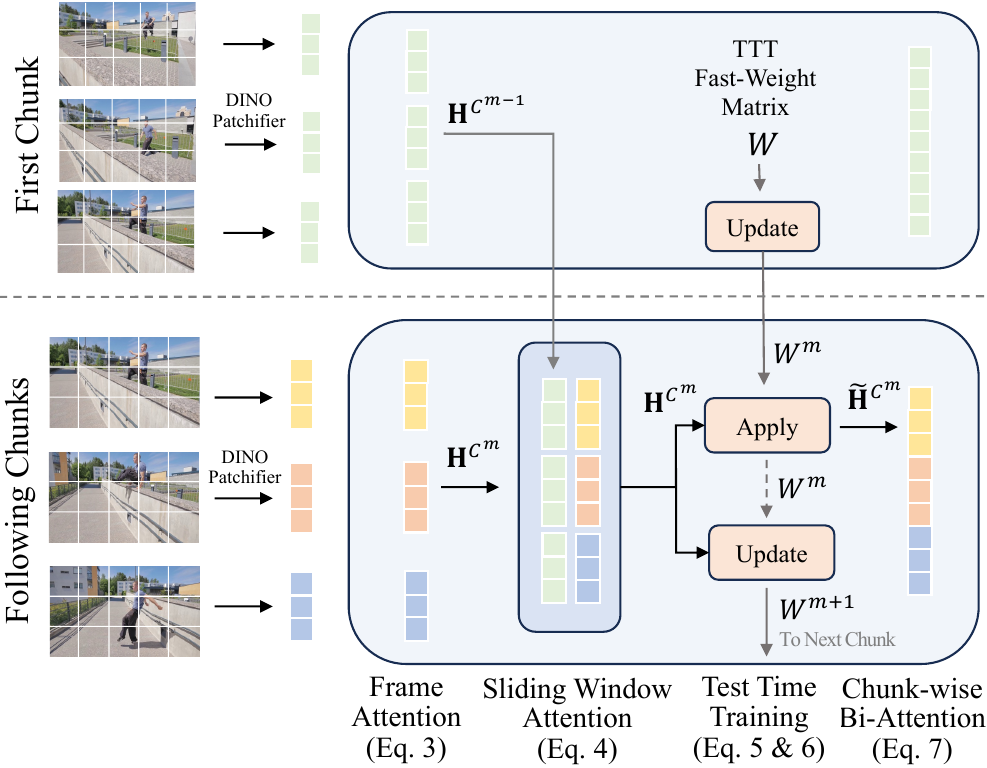}
    \vspace{-1.25em}
    \caption{\textbf{Overview of a single block of our hybrid memory module.} We process the input sequence in consecutive chunks of frames. While each block utilizes frame and bidirectional attention from prior work, we introduce new components to effectively propagate information across the entire sequence.
    Specifically, we incorporate Sliding Window Attention to improve consistency between neighboring chunks, and Test-Time Training layers to maintain long-range, global consistency across all chunks.
}
    \label{fig:system}
    \vspace{-0.5em}
\end{figure}
\vspace{-0.25em}
\section{Approach}
\subsection{Long-Context 3D Reconstruction Architecture}

\noindent \textbf{Motivation.}
To scale feedforward dense 3D reconstruction to minutes-long videos, we must overcome the quadratic complexity of global attention and the scarcity of long-horizon training data. End-to-end \textit{chunk-wise processing} emerges as the natural solution: it tightly bounds the computational cost and ensures that local inferences remain within the distribution of existing short-context training data. However, independently processing chunks inherently breaks global consistency. Therefore, we need a feedforward architecture that simultaneously provides:
(i) strong local bidirectional reasoning within a short chunk to preserve dense geometric fidelity,
(ii) a lossless short-range information highway to preserve high-precision geometric alignment across adjacent chunk boundaries, 
and (iii) a linear-time, fixed-size memory mechanism for long-range global propagation across thousands of frames.

\noindent \textbf{Overview.}
We process input video streams sequentially by chunk, as shown in~\cref{fig:teaser,fig:system}. 
Given a video $\mathcal{X}=\{I_t\}_{t=1}^{T}$, we partition it into $M$ chunks, $\{\mathcal{C}^m\}_{m=1}^{M}$, with minimal overlap (\eg, a single overlapping frame).
Each chunk $\mathcal{C}^m$ comprises a potentially variable number of frames $\mathcal{C}^m_j$, where $j$ indexes the frames within the chunk.
Within each chunk, we employ a strong bidirectional geometry backbone (\eg, VGGT or \pithree) to obtain high-quality dense predictions.
To propagate information across chunks, we introduce two complementary mechanisms:

\emph{(1) Long-term, lossy compression via chunk-wise TTT.}
We insert TTT layers that maintain a set of fast weights $W$ across a large number of chunks.
Aligning with our chunk-wise processing, we utilize Large-Chunk Test-Time Training (LaCT)~\cite{zhang2025test}, which has been shown to be more efficient than standard TTT.
During inference, the weights undergo both update and apply operations for each chunk.
In the apply operation, the TTT layers use historical information stored in the weights to modulate the network's processing of the current chunk. 
In the update operation, the weights are edited to store information from the current chunk, conceptually compressing important but redundant geometric information, \eg, coarse geometry and scale of the scene. 
While these fast weights theoretically offer an infinite receptive field, their practical capacity is inherently bounded by the training context length. To prevent drift over extremely long horizons, we optionally employ periodic state resets during inference (detailed in~\cref{sec:long_eval}).

\emph{(2) Short-term, lossless transfer via sliding-window attention (SWA).}
Relying solely on TTT-style state passing is inherently lossy, which is problematic for dense 3D reconstruction where geometric consistency across adjacent frames is critical.
To mitigate this, we \textit{sparsely insert} sliding-window attention layers that attend to the tokens output by the frame attention layer from both the previous and current chunks, \ie, $\mathcal{C}^{m-1} \cup \mathcal{C}^m$.
This establishes a lossless information highway that directly propagates high-fidelity features from the previous chunk.
Importantly, this operation remains efficient with bounded compute and memory, as sliding window attention is applied only between neighboring chunks and inserted at a subset of network depths (only four layers).

These two cross-chunk pathways are complementary: TTT provides scalable long-range memory, while SWA ensures fine-grained geometric consistency across adjacent chunks.

\noindent \textbf{Block structure.}
The geometry backbone first patchifies images into tokens and feeds them into a stack of residual network blocks.
In each network block, we introduce our hybrid memory system as shown in~\cref{fig:system}.
Denote slow weights (frozen at inference) by $\theta$.
Let $\mathcal{C}^m$ be a chunk of frames $\mathcal{C}^m_1, \ldots, \mathcal{C}^m_n$;   $\mathbf{H}^{C^m_i}$ be the token sequence specific to the frame $\mathcal{C}^m_i$; $\mathbf{H}^{\mathcal{C}^m}$ be the full token sequence entering a block for chunk $m$; and $\mathrm{LN}(\cdot)$ denote LayerNorm.
We list the detailed structure of one block below:

\emph{(1) Per-frame attention.} We apply self-attention independently to each frame's tokens to extract spatial features:
\vspace{-0.4em}
\begin{equation}
\mathbf{H}^{\mathcal{C}^{m}}
\leftarrow \mathbf{H}^{\mathcal{C}^{m}} + [\mathrm{Attn}_{\mathrm{frame}}\!\left(\mathrm{LN}(\mathbf{H}^{\mathcal{C}^m_i}) ;\theta\right),  | i\in \{1,\ldots,n\}]
\label{eq:frame_attn}
\end{equation}
where $[]$ is the concatenation operator.

\emph{(2) Sparse sliding-window attention (SWA) over $\mathcal{C}^{m-1}$ and $\mathcal{C}^{m}$.}
To align adjacent chunks, we insert SWA layers at a subset of depths (only 4 layers) to stay compute-bound:
\begin{equation}
\mathbf{H}^{\mathcal{C}^{m}}
\leftarrow \mathbf{H}^{\mathcal{C}^{m}} + \mathrm{Attn}_{\mathrm{swa}}\!\left([\mathrm{LN}(\mathbf{H}^{\mathcal{C}^{m-1}}),\mathrm{LN}( \mathbf{H}^{\mathcal{C}^{m}})];\,\theta\right).
\label{eq:swa}
\end{equation}
\emph{(3) Chunk-wise TTT layer (fast weights).}
To integrate global context, we maintain a set of fast weights $W^m$ that summarize information up to chunk $m$.
The TTT layer performs an \emph{apply-then-update} procedure at the chunk level.
We use pre-norm inside TTT to stabilize long-horizon streaming.
\begin{align}
\textbf{Apply:}&\quad
\tilde{\mathbf{H}}^{\mathcal{C}^{m}}
= \mathbf{H}^{\mathcal{C}^{m}} + f_{W^m}\!\left(\mathrm{LN}(\mathbf{H}^{\mathcal{C}^m})\right),
\label{eq:ttt_apply2}
\\[2pt]
\textbf{Update:}&\quad
W^{m+1}
= \mathcal{U}\!\left(W^m;\, {\mathbf{H}^{\mathcal{C}^{m}}}\right),
\label{eq:ttt_update2}
\end{align}

where $f_{W^m}(\cdot)$ is the fast-weight module (\eg, a SwiGLU MLP) parameterized by $W^m$,
and $\mathcal{U}(\cdot)$ denotes the online update rule (\eg, a gradient-based update with a self-supervised objective).
Intuitively, the \emph{apply} step injects the current memory into token representations, while the \emph{update} step writes the chunk's summary into $W$ for the next chunk.

\emph{(4) Chunk-wise bidirectional attention within chunk $\mathcal{C}^m$.}
Finally, within each chunk, we employ a bidirectional attention module for powerful geometric reasoning under a bounded context window. The updated representation $\mathbf{H}^{\mathcal{C}^{m}}$ then serves as the input to the subsequent network block:
\begin{equation}
\mathbf{H}^{\mathcal{C}^{m}}
\leftarrow \tilde{\mathbf{H}}^{\mathcal{C}^{m}} + \mathrm{BiAttn}_{\mathrm{chunk}}\!\left(\mathrm{LN}(\tilde{\mathbf{H}}^{\mathcal{C}^{m}});\,\theta\right).
\label{eq:chunk_biattn}
\end{equation}
\emph{Prediction heads.}
After the final residual block, we attach lightweight decoders---a pointmap decoder and a camera-pose decoder---following $\pi^3$, to produce the final dense pointmap predictions and per-frame camera poses.

\begin{figure}[t] %
    \centering
        \begin{subfigure}[b]{0.5\textwidth} %
        \includegraphics[width=\linewidth]{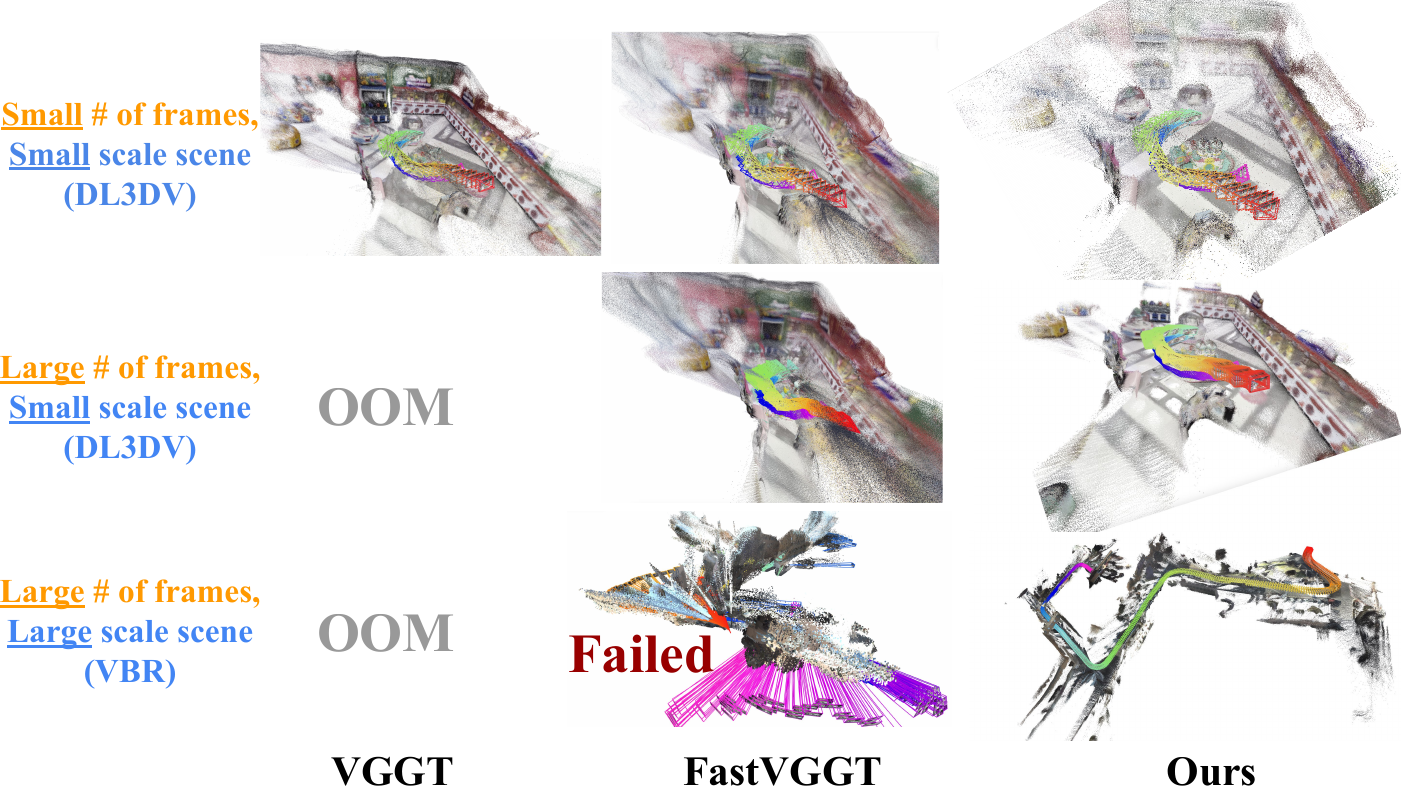} %
    \end{subfigure}
    \caption{\textbf{Comparison of different methods across varying sequence lengths and scene scales.} Although FastVGGT is able to process a larger number of frames during inference, it fails completely on large-scale scenes, highlighting the inherent ``data wall'' of models trained strictly on short-context bubbles. In contrast, \ourmethod{} breaks both the context and data walls by pairing a hybrid memory architecture with diverse long-horizon training data.
}
    \label{fig:data_scenescale}
    \vspace{-1em}
\end{figure}

\begin{table*}[t]
\centering
\caption{\textbf{Comparison of Absolute Trajectory Error (ATE$\downarrow$, m) on KITTI.} The top and bottom blocks denote \textit{optimization-based} and \textit{feedforward} methods, respectively. \textbf{Bold} and \underline{underline} indicate the best and second-best performance among feedforward methods. \ourmethod{}* denotes a variant of our method with feedforward pose alignment for both training and inference.
}
\vspace{-0.5em}
\label{tbl:kitti}

\resizebox{\textwidth}{!}{%

\begin{tabular}{clccccccccccc|c}
\toprule
& \textbf{Methods} & 00 & 01 & 02 & 03 & 04 & 05 & 06 & 07 & 08 & 09 & 10 & \textbf{Avg.} \\
& \scriptsize \textit{num. of frames, scale} & \scriptsize 4542, 3.7km & \scriptsize 1101, 2.5km & \scriptsize 4661, 5.1km & \scriptsize 801, 0.6km & \scriptsize 271, 0.4km & \scriptsize 2761, 2.2km & \scriptsize 1101, 1.2km & \scriptsize 1101, 0.7km & \scriptsize 4071, 3.2km & \scriptsize 1591, 1.7km & \scriptsize 1201, 0.9km & - \\
& \scriptsize \textit{Contains Loop?} & \checkmark & $\times$ & \checkmark & $\times$ & $\times$ & \checkmark & \checkmark & \checkmark & $\times$ & \checkmark & $\times$ & - \\
\midrule
\multirow{4}{*}{\rotatebox{90}{\textbf{Opt.-based}}} 
& DROID-SLAM~\citep{teed2021droid} & 92.10 & 344.60 & 107.61 & 2.38 & 1.00 & 118.50 & 62.47 & 21.78 & 161.60 & 72.32 & 118.70 & 100.28 \\
& DPV-SLAM~\citep{lipson2024deep} & 112.80 & 11.50 & 123.53 & 2.50 & 0.81 & 57.80 & 54.86 & 18.77 & 110.49 & 76.66 & 13.65 & 53.03 \\
& DPV-SLAM++~\citep{lipson2024deep} & 8.30 & 11.86 & 39.64 & 2.50 & 0.78 & 5.74 & 11.60 & 1.52 & 110.90 & 76.70 & 13.70 & 25.75 \\
& VGGT-Long~\cite{deng2025vggt} & 8.67 & 121.17 & 32.08 & 6.12 & 4.23 & 8.31 & 5.34 & 4.63 & 53.10 & 41.99 & 18.37 & 27.64 \\
\midrule
\multirow{7}{*}{\rotatebox{90}{\textbf{Feedforward}}} 
& FastVGGT~\cite{shen2025fastvggt} & OOM & 639.39 & OOM & 21.53 & 9.51 & OOM & 40.56 & 51.35 & OOM & 201.54 & 196.22 & -- \\
& InfiniteVGGT~\cite{yuan2026infinitevggt} & 186.46 & 623.62 & 289.16 & 166.74 & 68.00 & 143.84 & 117.57 & 85.33 & 221.56 & 215.41 & 156.92 & 206.78 \\
& CUT3R~\citep{wang2025continuous} & 190.38 & 90.59 & 264.39 & 20.40 & 7.31 & 92.25 & 67.54 & 22.48 & 145.08 & 67.42 & 40.00 & 91.62 \\
& TTT3R~\citep{chen2025ttt3r} & 119.94 & 99.59 & 238.07 & 16.83 & 3.98 & 36.38 & 47.20 & 11.62 & 107.33 & 86.96 & 33.58 & 72.86 \\
\cmidrule{2-14}
& Pi3-Chunk (Proposed Baseline) & \textbf{26.65} & 196.04 & 157.92 & \underline{5.13} & \textbf{1.09} & \textbf{12.79} & 27.66 & \underline{5.94} & 61.26 & 56.31 & 21.96 & 52.07 \\
& \cellcolor{gray!10}\textbf{\ourmethod{}} \textbf{(Ours)} & \cellcolor{gray!10}62.34 & \cellcolor{gray!10}\textbf{41.64} & \cellcolor{gray!10}\underline{39.64} & \cellcolor{gray!10}\textbf{4.89} & \cellcolor{gray!10}\underline{1.82} & \cellcolor{gray!10}41.27 & \cellcolor{gray!10}\underline{13.99} & \cellcolor{gray!10}16.24 & \cellcolor{gray!10}\underline{26.46} & \cellcolor{gray!10}\underline{22.71} & \cellcolor{gray!10}\textbf{8.84} & \cellcolor{gray!10}\underline{25.44} \\
& \cellcolor{gray!10}\textbf{\ourmethod{}*} \textbf{(Ours)} & \cellcolor{gray!10}\underline{30.47} & \cellcolor{gray!10}\underline{47.91} & \cellcolor{gray!10}\textbf{36.32} & \cellcolor{gray!10}5.38 & \cellcolor{gray!10}1.95 & \cellcolor{gray!10}\underline{26.34} & \cellcolor{gray!10}\textbf{6.60} & \cellcolor{gray!10}\textbf{5.55} & \cellcolor{gray!10}\textbf{24.41} & \cellcolor{gray!10}\textbf{10.12} & \cellcolor{gray!10}\underline{10.11} & \cellcolor{gray!10}\textbf{18.65} \\
\bottomrule
\end{tabular}%

}
\vspace{-1.5em}
\end{table*}

\subsection{Learning Objectives}
\label{sec:training}

\noindent \textbf{Objective.}
We follow $\pi^3$~\citep{wang2025pi} and train \ourmethod{} with (i) a \emph{scale-invariant local pointmap loss} and (ii) an \emph{affine-invariant relative pose loss}, where the losses do not require a reference view.
We align the predicted local pointmaps with a single per-sequence scale $s^{\ast}$ as in MoGe~\citep{wang2025moge}, and compute a reconstruction loss, normalized by depth values, over all pixels.
To further over-constrain long-sequence training, we additionally impose a \emph{global pointmap loss} on world-coordinate pointmaps obtained by transforming local pointmaps using the predicted camera poses.
\begin{align}
\mathcal{L}_{\text{local}}
&\!=\! \frac{1}{N|\Omega|}\sum_{i=1}^{N}\sum_{p\in\Omega}
\frac{1}{z_{i,p}}\,\big\|\, s^{\ast}\hat{\mathbf{x}}_{i,p}-\mathbf{x}_{i,p}\,\big\|_{1},
\qquad
\label{eq:local_pointmap_simple}
\\
\mathcal{L}_{\text{pose}}
&\!=\! \sum_{(i,j)\in\mathcal{P}}
\Big(
\lambda_{\text{r}}\mathcal{L}_{\text{rot}}(\hat{\mathbf{R}}_{ij}, \mathbf{R}_{ij})
\!+\!
\lambda_{\text{t}}\big\|\, s^{\ast}\hat{\mathbf{t}}_{ij}\!-\!\mathbf{t}_{ij}\,\big\|_{\text{Huber}}
\Big),
\label{eq:pose_loss_simple}
\\
\mathcal{L}_{\text{global}}
&\!=\! \frac{1}{N|\Omega|}\sum_{i=1}^{N}\sum_{p\in\Omega}
\big\|\,
s^{\ast}\Pi(\hat{\mathbf{T}}_{i},\,\hat{\mathbf{x}}_{i,p})
\!-\!
\Pi(\mathbf{T}_{i},\,\mathbf{x}_{i,p})
\,\big\|_{1},
\label{eq:global_pointmap_simple}
\\
\mathcal{L}&\!=\! \mathcal{L}_{\text{local}}+\mathcal{L}_{\text{pose}}+\lambda_{\text{global}}\mathcal{L}_{\text{global}}.
\end{align}
Here, $N$ is the number of frames used in the supervision set, $\Omega$ is the pixel index set (so $|\Omega|=HW$),
$\hat{\mathbf{x}}_{i,p}\in\mathbb{R}^3$ and $\mathbf{x}_{i,p}\in\mathbb{R}^3$ denote predicted and ground-truth local point coordinates at pixel $p$ in frame $i$,
and $z_{i,p}$ is the corresponding depth for normalization.
$\hat{\mathbf{T}}_i=[\hat{\mathbf{R}}_i\mid \hat{\mathbf{t}}_i]\in\mathrm{SE}(3)$ is the predicted camera pose, and
$\hat{\mathbf{R}}_{ij},\hat{\mathbf{t}}_{ij}$ denote the predicted relative motion between frames $i$ and $j$ (defined analogously for ground truth).
$\mathcal{P}$ is the set of supervised frame pairs (\eg, pairs within a chunk and/or overlap pairs across different chunks).
$\Pi(\mathbf{T},\mathbf{x})$ maps a local point $\mathbf{x}$ to world coordinates using pose $\mathbf{T}$.

\noindent \textbf{\ourmethod{}$^{\ast}$ feedforward alignment.}
Despite having TTT and SWA, very long streams may still accumulate prediction errors. To mitigate this, we introduce \ourmethod{}$^{\ast}$, a variant that incorporates a purely feedforward alignment step to align raw predictions into a consistent global coordinate system.
Let $\hat{\mathbf{T}}^{(m)}_{k}$ denote the raw predicted pose of the \emph{overlapping frame} $k$ within the current chunk $\mathcal{C}^m$, and let $\tilde{\mathbf{T}}^{(m-1)}$ denote the aligned poses of chunk $\mathcal{C}^{m-1}$ (initialized as $\tilde{\mathbf{T}}^{(1)}=\hat{\mathbf{T}}^{(1)}$).
We compute a rigid SE(3) alignment $\mathbf{A}_m$ that maps the current chunk $\mathcal{C}^m$ to the aligned coordinate system of the previous chunk $\mathcal{C}^{m-1}$ using the overlap: $\mathbf{A}_m
= \tilde{\mathbf{T}}^{(m-1)}_{k}\big(\hat{\mathbf{T}}^{(m)}_{k}\big)^{-1}$. This transformation is applied to all frames in $\mathcal{C}^m$:
\begin{align}
\tilde{\mathbf{T}}^{(m)}_{t}
= \mathbf{A}_m\,\hat{\mathbf{T}}^{(m)}_{t},\ \ \forall t\in\mathcal{C}^m .
\label{eq:loger_star}
\end{align}
We use $\tilde{\mathbf{T}}^{(m)}_{t}$ as the final camera pose prediction of \ourmethod{}$^{\ast}$ for \textit{both} training and inference.

\begin{figure}[t] %
    \centering
        \begin{subfigure}[b]{0.425\textwidth} %
        \includegraphics[width=\linewidth]{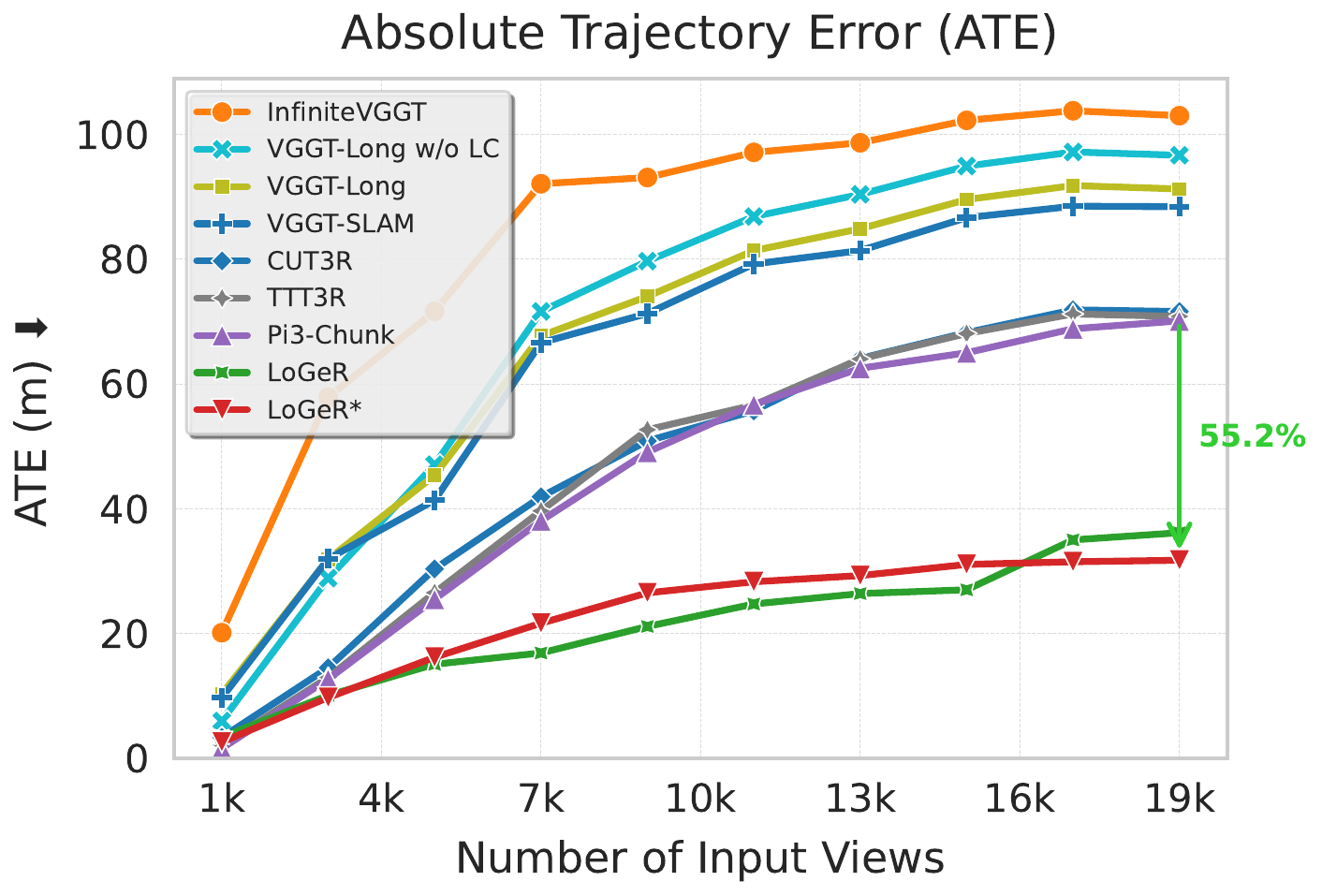} %
    \end{subfigure}
    \vspace{-0.5em}
    \caption{\textbf{Quantitative results on our proposed VBR}~\cite{brizi2024vbr} evaluation showing results on very long sequences spanning from 1,000 to 19,000 frames. 
    Our methods achieve 55.2\% more accurate results than prior methods.
    }
    \vspace{-1em}
    \label{fig:quant_vbr}
\end{figure}

\subsection{Data and Curriculum}
\label{sec:data}

\noindent \textbf{The ``Data Wall".}
We posit that architectural improvements alone are insufficient for infinite-context reconstruction. 
As shown in~\cref{fig:data_scenescale}, we observe that strong baselines like VGGT~\cite{wang2025vggt}, even when equipped with inference-time architectural efficiency improvement (FastVGGT~\citep{shen2025fastvggt}), fail to generalize to large-scale scenes when trained solely on short-context or small-scene-scale data. 
To overcome this ``data wall'', we construct a training mixture heavily weighted towards large-scale-scene datasets, \eg, TartanAirV2~\cite{patel2025tartanground}, which provides the necessary long-horizon signal for learning effective geometry compression. 

\noindent \textbf{Curriculum Training.}
Crucially, to stabilize the optimization of recurrent TTT layers, we employ a \textbf{progressive curriculum strategy}.
By starting with easier sequences and progressively increasing complexity, we force the model to shift reliance from local Sliding Window Attention to the global TTT hidden state.
Our schedule proceeds in three stages: 
(1) we begin with 48 frames split into 4 chunks; 
(2) we {gradually increase chunk density} to 12 chunks while maintaining fixed sequence length; 
and (3) leveraging H200 GPUs, we scale the context length to 128 frames, progressively increasing to 20 chunks.
For \ourmethod{}$^{\ast}$, we initialize from the first-stage model, integrate the feedforward alignment, and fine-tune through the remaining curriculum.
This strategy not only improves training efficiency by reducing train-time rollout overhead, but also boosts final performance, as in the ablation study in \cref{subsec:ablation_study,sec:discussion}.

\begin{figure*}[t] %
    \centering
    \vspace{-0.5em}
        \begin{subfigure}[b]{0.975\textwidth} %
        \includegraphics[width=\linewidth]{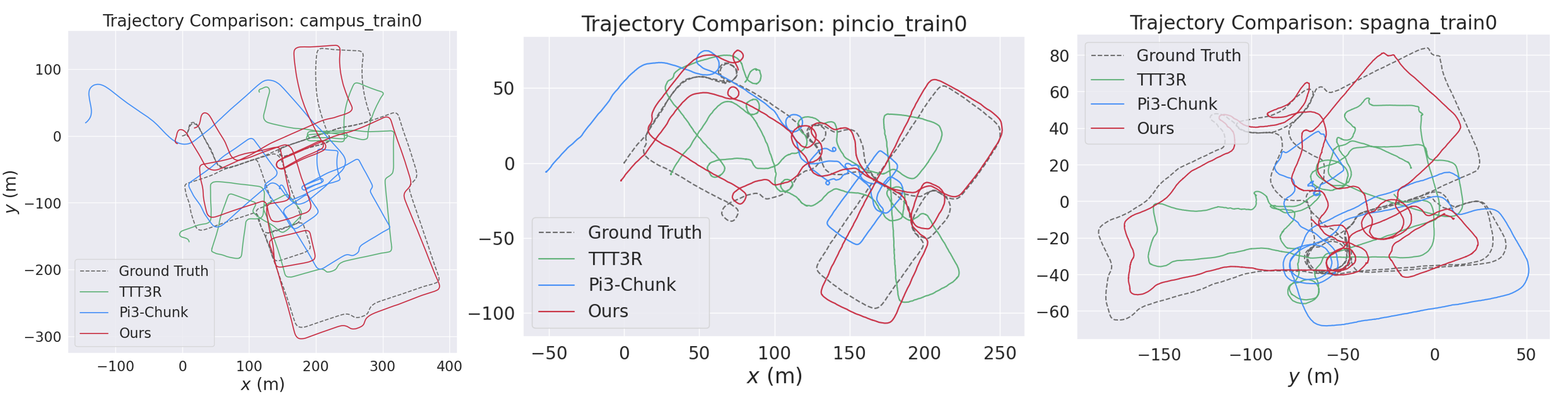} %
    \end{subfigure}
    \vspace{-1em}
\caption{\textbf{Qualitative camera trajectories on the VBR dataset.} \ourmethod{} accurately preserves global scale and trajectory over very long sequences, closely matching the ground truth where prior methods suffer from severe drift.}
    \label{fig:qual_vbr}
    \vspace{-0.25em}
\end{figure*}

\section{Experiments}

\begin{figure*}[t!]
    \centering
    \vspace{-0.5em}
    \begin{minipage}[h]{0.35\textwidth}
        \centering
        \includegraphics[width=\linewidth]{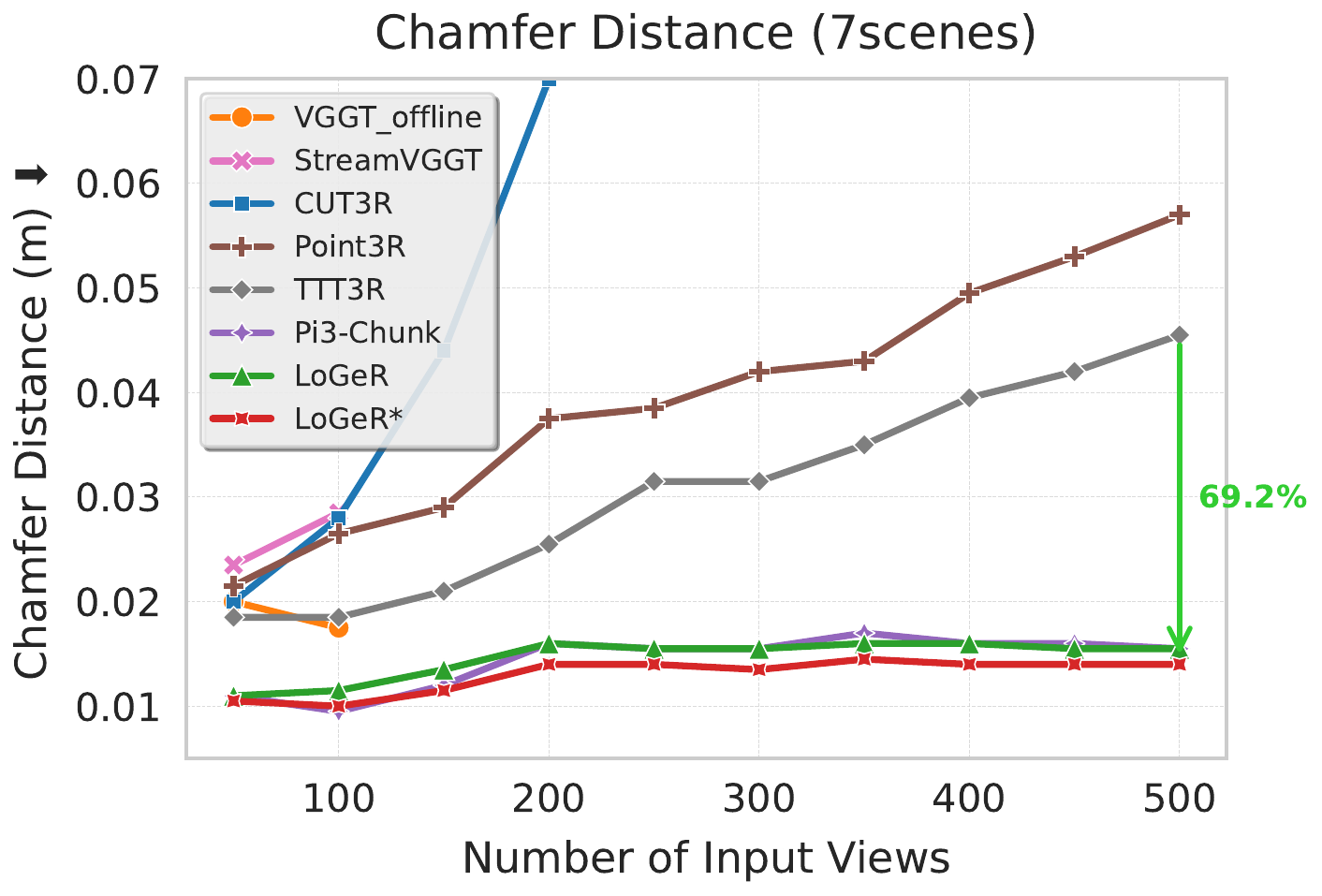}
        \vspace{-1.75em}
        \caption{\textbf{3D reconstruction result on 7scenes.} Both our proposed baseline and \ourmethod{} significantly outperform prior work by \textbf{69.2\%}.}
        \label{fig:quant_7scenes}
    \end{minipage}
    \hfill
    \begin{minipage}[h]{0.62\textwidth}
        \centering
        % \vspace{-0.5em}
        \includegraphics[width=\linewidth]{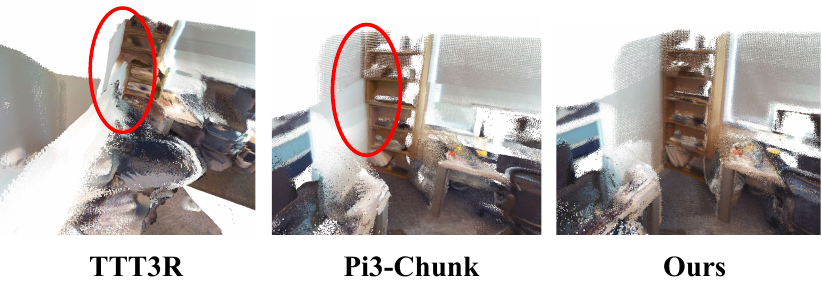}
        \vspace{-1.75em}
        \caption{\textbf{Qualitative results on 3D reconstruction for the 7scenes dataset} demonstrating the \ourmethod{} outperforms both our baseline and prior work TTT3R. \ourmethod{} accurately reconstructs the bookshelf, while both our proposed strong baseline and TTT3R cause large distortions.
        }
        \label{fig:qual_7scenes}
    \end{minipage}
    \vspace{-1.25em}
\end{figure*}

\noindent \textbf{Implementation details.}
We train our model on a mixture of real and synthetic datasets, including
ARKitScenes~\cite{baruch2021arkitscenes},
DL3DV~\citep{ling2024dl3dv},
HyperSim~\cite{roberts2021hypersim},
MegaDepth~\cite{li2018megadepth},
ScanNet~\cite{dai2017scannet},
ScanNet++~\cite{yeshwanth2023scannet++},
Spring~\cite{mehl2023spring},
TartanAir~\citep{wang2020tartanair},
TartanAirV2~\citep{patel2025tartanground},
UnReal4K~\cite{wang2023neural},
Virtual KITTI 2~\citep{cabon2020vkitti2},
Waymo~\cite{schwall2020waymo},
and a subset of OmniWorld-Game.
We optimize the model with AdamW~\citep{loshchilov2017decoupled} for 40k steps with a batch size of 32. The training process requires approximately two days on 32 NVIDIA H100 GPUs, followed by another two days on 32 H200 GPUs.
We initialize the weights of the patchifier, frame attention, and chunk-wise bidirectional attention modules from $\pi^3$~\citep{wang2025pi}. All the evaluation is done on a NVIDIA A100 40GB. See more details in Appendix~\cref{subsec:supp_data_details,subsec:more_training_details,subsec:more_arch_details}.

\begin{figure*}[t!] %
    \centering
    \vspace{-0.5em}
        \includegraphics[width=0.875\linewidth]{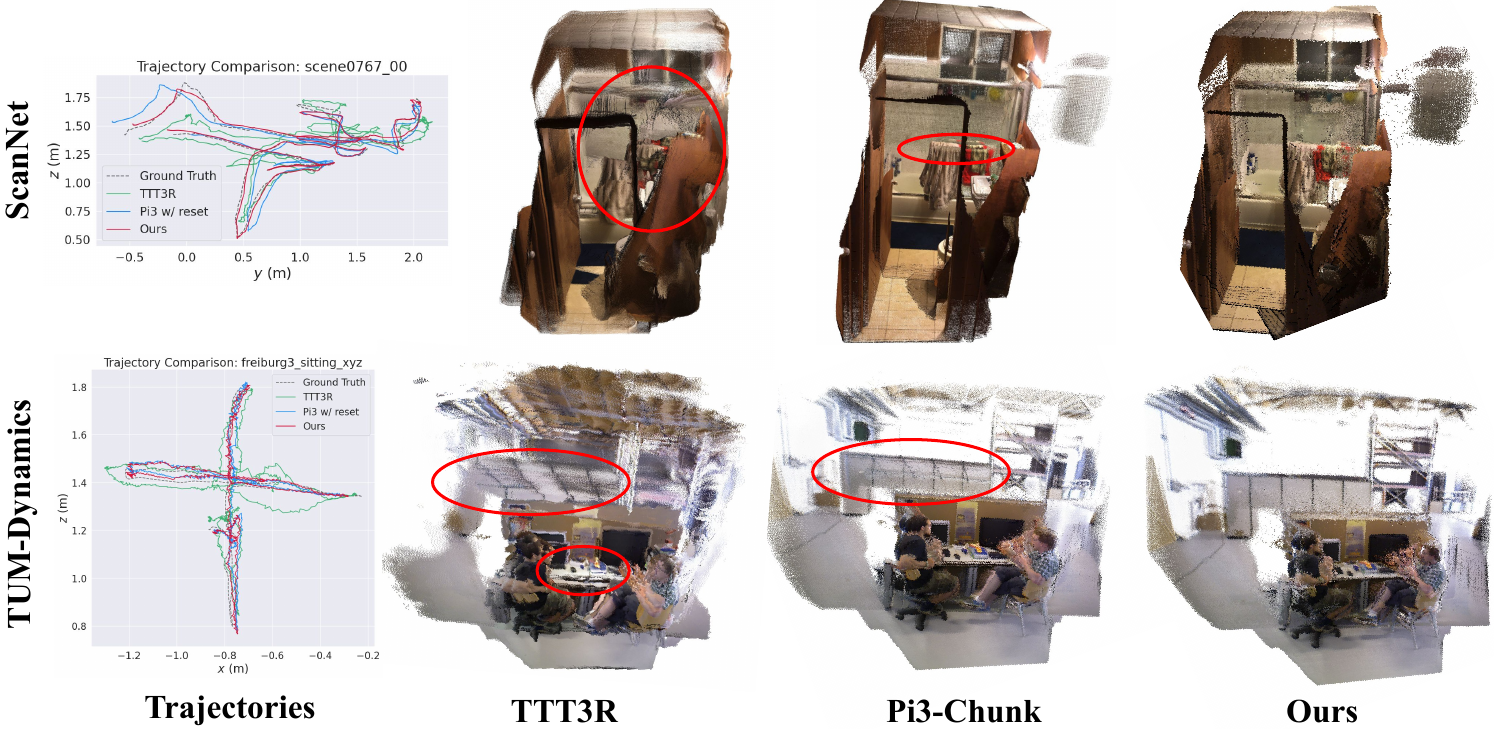} %
        \vspace{-0.25em}
    \caption{\textbf{Qualitative results on ScanNet and TUM-Dynamics.} While the proposed Pi3-Chunk baseline yields slightly better pose metrics on small-scale TUM sequences (unlike its severe drift on longer trajectories), \ourmethod{} produces visually superior reconstructions. 
    As highlighted, \ourmethod{} accurately recovers structural details, avoiding the severe distortions and geometric artifacts present in baselines.}
    \label{fig:qual_scannettum}
\end{figure*}

\begin{figure*}[t!] %
    \centering
    \vspace{-0.5em}
        \begin{subfigure}[b]{0.8\textwidth} %
        \includegraphics[width=\linewidth]{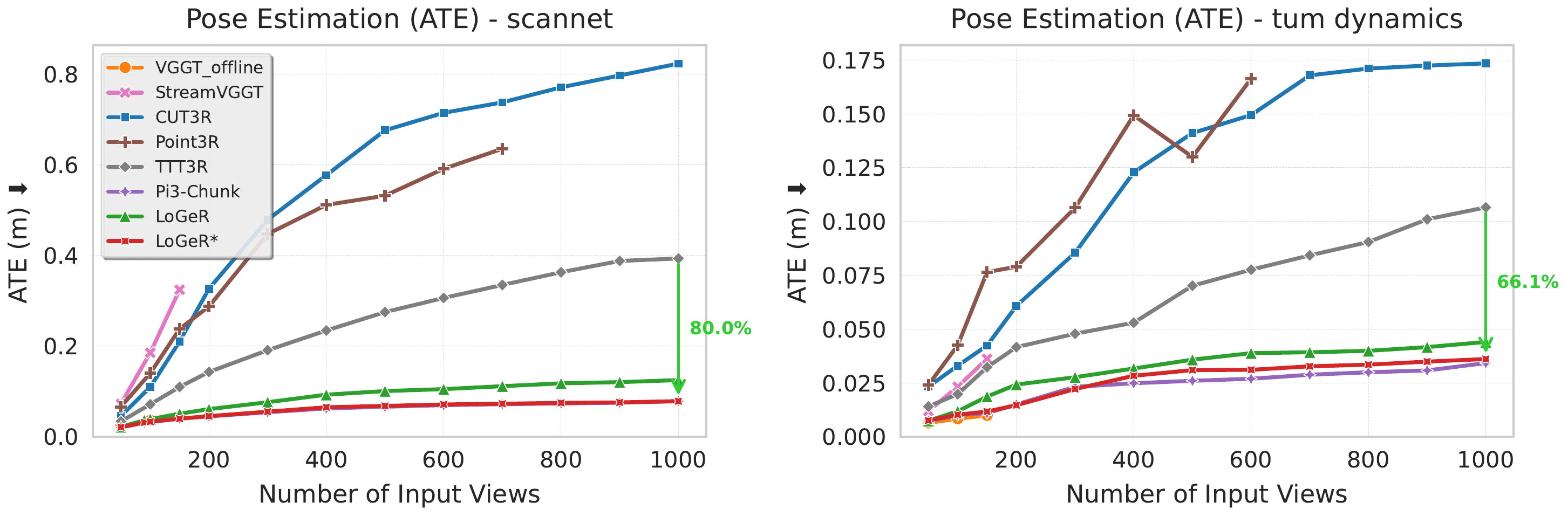} %
    \end{subfigure}
    
    \vspace{-0.5em}
    \caption{Q\textbf{uantitative results on camera pose evaluation}, demonstrating that both our proposed baseline and \ourmethod{} substantially outperform prior work, with \textbf{80.0\%} and \textbf{66.1\%} relative gains on ScanNet and TUM datasets, respectively.}
    \vspace{-1.5em}
    \label{fig:quant_scannettum}
\end{figure*}

\subsection{Evaluation on Long Sequences}
\label{sec:long_eval}

\noindent\textbf{Benchmarks.} 
As shown in~\cref{fig:data_scenescale}, for large-scale geometric reconstruction, evaluation must be conducted on expansive scenes. To this end, we evaluate our method on two long-sequence benchmarks. First, we use the standard \textbf{KITTI benchmark}~\citep{geiger2012we}, which contains sequences up to 4,661 frames and trajectory lengths up to 5 km. To further challenge the robustness of our method on even more complex and longer trajectories, we identify and repurpose the \textbf{VBR dataset}~\citep{brizi2024vbr}. This dataset consists of video sequences taken from Rome with paired ground truth 3D reconstructions derived from refined LiDAR point clouds and bundle-adjusted camera poses. Specifically, we use 7 sequences ranging from 8,815 to 18,846 frames and covering distances between 1.4 km and 11.5 km. We report the Absolute Trajectory Error (ATE) of the predicted trajectory with respect to the ground truth after Umeyama alignment~\cite{umeyama1991least}, following standard protocols~\cite{teed2021droid, lipson2024deep}.

\noindent\textbf{Baselines.} 
We conduct a comprehensive comparison against both optimization-based and feedforward approaches. 
For \textit{optimization-based} methods, we compare with classic dense SLAM systems such as DROID-SLAM~\cite{teed2021droid} and DPV-SLAM~\cite{lipson2024deep}, as well as recent foundation-model-based approaches like VGGT-SLAM and VGGT-Long~\cite{wang2025vggt}.
For \textit{feedforward} methods, we evaluate recurrent architectures including CUT3R~\cite{wang2025continuous} and TTT3R~\cite{chen2025ttt3r}, alongside bidirectional models such as FastVGGT~\cite{shen2025fastvggt} and autoregressive approaches like InfiniteVGGT~\cite{yuan2026infinitevggt}.
For CUT3R and TTT3R, we adopt the reset algorithm as proposed in TTT3R to obtain a reasonable result.
For our method, we also reset the fast weights in the TTT layers after every five windows to avoid error accumulation within a fixed size of state~\citep{ruiz2025understanding}. Note, we also apply the feedforward pose alignment when doing a reset. See Appendix~\cref{subsec:more_inference_details} for more details.
Based on the philosophy of chunk-causal, we also propose a simple baseline built on top of $\pi^3$ for long sequence reconstruction. Specifically, we process the input in a chunk-wise manner as in \ourmethod{}, and then compute a SIM(3) transformation based on the overlapping frames to stitch the predictions of different chunks together. We name this new baseline \textit{Pi3-Chunk}. See more details in Appendix~\cref{subsec:more_baseline_details}.

\noindent\textbf{Results.} 
Quantitatively, both \ourmethod{} and our proposed baseline, Pi3-Chunk, significantly outperform existing feedforward methods on the \textbf{KITTI benchmark} (\cref{tbl:kitti}). Notably, \ourmethod{} achieves an average performance that surpasses even the strongest optimization-based method, VGGT-Long, by $32.5\%$. This advantage is particularly evident on open-loop trajectories (\ie, 01, 03, 04, 08, and 10), where our method effectively mitigates accumulated drift without relying on loop closure.
On the \textbf{VBR benchmark}, consistent improvements are observed quantitatively in~\cref{fig:quant_vbr} and qualitatively in~\cref{fig:qual_vbr}. 
While Pi3-Chunk yields slightly better results at shorter horizons (\eg, 1k frames), the advantage of \ourmethod{} becomes increasingly prominent as the sequence length grows. 
This is because Pi3-Chunk relies on local overlapping frames for $\text{SIM}(3)$ scale estimation, causing scale errors to accumulate exponentially over extended distances. In contrast, \ourmethod{}'s TTT module inherently anchors the global scale.
This is further evidenced by qualitative visualizations, where \ourmethod{} maintains global scale consistency up to 20k frames, whereas the baseline exhibits severe scale drift in such extremely long sequences. See Appendix~\cref{subsec:more_vbr,subsec:more_kitti,subsec:gallery} for more long sequence results.

\begin{figure*}[t]
    \centering

    \includegraphics[width=0.9125\linewidth]{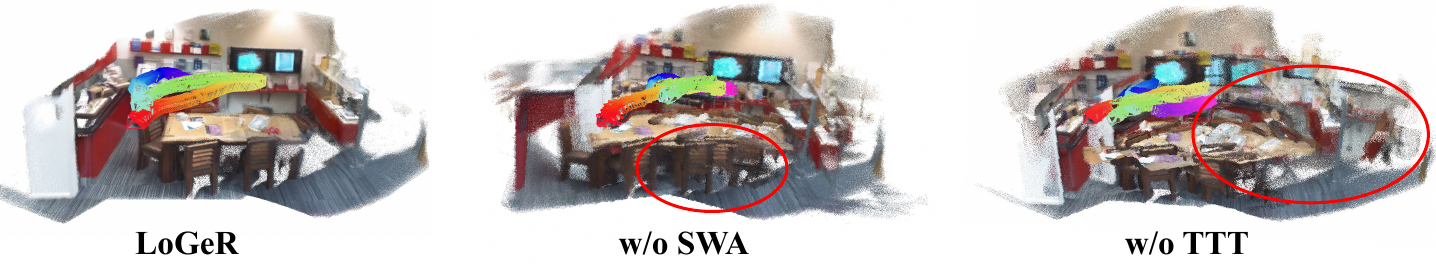}
    \vspace{-0.5em}

    \caption{\textbf{Ablation of disabling SWA or TTT at inference time.} SWA ensures geometric consistency between adjacent chunks (local consistency), and lacking it causes noticeable local misalignment artifacts. Meanwhile, TTT maintains the global context (global consistency); without it, the model suffers from severe trajectory drift over long horizons.}
    \label{fig:ablation_inference}
    \vspace{-1.15em}
\end{figure*}

\subsection{Evaluation on Short Sequences}
Following TTT3R, we extend our evaluation to shorter sequences (up to 1k frames).
First, we evaluate 3D point-cloud reconstruction on 7-Scenes~\cite{shotton2013scene} with sequence lengths ranging from 50 to 500 frames.
Subsequently, we evaluate camera pose estimation on ScanNetV2~\cite{dai2017scannet} and TUM-Dynamics~\cite{tum} for sequence lengths ranging from 50 to 1k frames.
For all experiments in this section, we use a chunk size of 64 and an overlap of 3 frames.
We compare our approach with other learning-based sub-quadratic baselines, including explicit-state methods such as Point3R and implicit state-space models such as CUT3R, TTT3R, and StreamVGGT. Additionally, we include comparisons with bidirectional baselines, specifically VGGT and $\pi^3$.
Results for 7-Scenes are presented in~\cref{fig:quant_7scenes,fig:qual_7scenes}, while results for ScanNetV2 and TUM-Dynamics are shown in~\cref{fig:qual_scannettum,fig:quant_scannettum}. 
Across all comparisons, both our proposed baseline and \ourmethod{} significantly outperform prior work.
See Appendix~\cref{subsec:7scenes_eval,subsec:depth_eval} for more evaluation on short sequences.

\subsection{Ablation Study}
\label{subsec:ablation_study}

In this section, we validate the effectiveness of our proposed hybrid architecture, data mixture, and training curriculum. 
For computational efficiency, all ablation models are trained with a reduced number of frames compared to the final model and evaluated on a subset of ScanNet and the TUM-Dynamics dataset. 
Quantitative results are in~\cref{tab:experiment_results}.

\noindent \textbf{Architecture Design.}
First, we assess the contribution of the two core memory components: the TTT layer and the SWA layer. 
As shown in the first block of~\cref{tab:experiment_results}, removing either component leads to a noticeable performance drop in ATE, confirming that the hybrid design is essential for robust state estimation.
To further intuit the role of each module, we provide a qualitative comparison in~\cref{fig:ablation_inference}, where we selectively disable the SWA or TTT layers of a trained \ourmethod{} model at \textit{inference time}. 
Visualizing the trajectory demonstrates that both components are critical: TTT ensures global consistency, while SWA maintains local smoothness, and lacking either results in significant drift or geometric degradation.

\noindent \textbf{Dataset Mixture.}
Next, we verify the necessity of including large-scale navigation datasets to overcome the ``data wall''. 
We train a variant of our model by excluding five specific large-scale datasets: TartanAir, TartanAirV2, Waymo, Virtual KITTI 2, and OmniWorld-Game.
The results in the second block of~\cref{tab:experiment_results} show that the model trained on the full data mixture significantly outperforms the one trained without these large-scale sources, validating our hypothesis that diverse, long-horizon data is key to generalization.

\noindent \textbf{Curriculum Training.}
Finally, we analyze the impact of our progressive curriculum strategy. 
We compare the standard training schedule against our curriculum approach for both \ourmethod{} and \ourmethod{}*. 
As reported in the bottom blocks of~\cref{tab:experiment_results}, the curriculum strategy consistently improves performance across both model variants and datasets, demonstrating its effectiveness in stabilizing the optimization of recurrent layers and feedforward alignment.

\begin{table}[t]
\footnotesize
\centering
\caption{\textbf{Comparison of baseline and ablation results} across ScanNet (subset) and TUM datasets. We report the results of ATE ($\downarrow$) on both 500 and 1000 frames of sequences.}
\vspace{-0.25em}
\begin{tabular}{lcccc}
\toprule
& \multicolumn{2}{c}{\textbf{ScanNet}} & \multicolumn{2}{c}{\textbf{TUM}} \\
\cmidrule(lr){2-3} \cmidrule(lr){4-5}
\textbf{Method} & 500f & 1000f & 500f & 1000f \\
\midrule
\textbf{\ourmethod{}} & \textbf{0.087} & \textbf{0.107} & \textbf{0.033} & \textbf{0.050} \\
w/o TTT & 0.108 & 0.162 & 0.043 & 0.079 \\
w/o SWA & 0.115 & 0.143 & 0.039 & 0.053 \\
\midrule
\textbf{All datasets} & \textbf{0.087} & \textbf{0.107} & \textbf{0.033} & \textbf{0.050} \\
w/o 5 large datasets & 0.102 & 0.156 & 0.050 & 0.072 \\
\midrule
\textbf{\ourmethod{}} & \textbf{0.087} & \textbf{0.107} & \textbf{0.033} & \textbf{0.050} \\
w/o curriculum & 0.098 & 0.133 & 0.049 & 0.062 \\
\midrule
\textbf{\ourmethod{}*} & \textbf{0.070} & \textbf{0.080} & 0.031 & \textbf{0.036} \\
w/o curriculum & 0.078 & 0.093 & \textbf{0.029} & 0.040 \\
\bottomrule
\end{tabular}
\label{tab:experiment_results}
\vspace{-1.75em}
\end{table}

\section{Conclusion}

We addressed the limitations of long-sequence estimation for feedforward 3D geometry models by introducing \ourmethod{}, a complementary hybrid architecture combining Sliding Window Attention (SWA) with Test-Time Training (TTT) layer.
Validated on our benchmark, \ourmethod{} processes thousands of frames without optimization, outperforming other methods on long-sequence reconstruction with various representations, \eg, explicit memories, recurrent states, or causal attention.
Our work opens new avenues for long-context spatio-temporal reasoning in dynamic scenes, video understanding, and robotics.

\section*{Discussion and Future work}

\label{sec:discussion}

Our findings open several promising future directions.
First, while TTT fast weights have a fixed memory footprint that theoretically allows infinite context, in practice they struggle to generalize beyond the number of chunks they were trained with~\citep{ruiz2025understanding}, restricting their effective range to the training context length (which is constrained by hardware memory budgets). Exceeding this on extremely long sequences (\eg $>1,000$ frames) causes error accumulation and trajectory drift. Preventing this currently requires periodic state resets that sacrifice long-term context. We hope future linear sequence models will resolve this length-generalization bottleneck.
Second, the availability of expansive, high-quality training data remains a significant bottleneck, described as the ``data wall" discussed in~\cref{sec:data}. We hope future community efforts will focus on curating more diverse, long-horizon datasets. 
Finally, we aim to extend our hybrid memory architecture (SWA + TTT) beyond geometric reconstruction, exploring its potential in other domains that demand both long-term global consistency and strong local dependencies.

\section*{Acknowledgment}
We would especially like to thank Noah Snavely for helpful feedback throughout the project. We thank Tianyuan Zhang, Songlin Yang, Youming Deng, Haiwen Feng, Qianqian Wang, and Yifei Zhang for helpful discussions. 
We thank Alfred Piccioni for the help with the training infrastructure. 
We thank Xingyu Chen for providing details on the evaluation. 
We thank Zihan Zhu for his feedback on the evaluation of the VBR benchmark.
We thank Angjoo Kanazawa, Tyler Bonnen, and Jiahui Lei for helpful feedback on the manuscript.

\section*{Impact Statement}

This paper presents work whose goal is to address the problem of long-context 3D reconstruction design using a new architectural component. The main societal consequences of our work come from any potential improvements in the ability to convert video into dense 3D scenes, such as improved applications in VR, generative 3D, and Robotics. None of these consequences is specific to our work and applies to all techniques that improve the task of 3D reconstruction or design new architectural components for deep learning architectures.

\bibliography{example_paper}
\bibliographystyle{icml2026}

\clearpage
\newpage
\appendix
\section*{Appendix}
\vspace{0.5em}
\section*{Content}

\noindent \textbf{A. Additional Implementation Details} \dotfill \pageref{sec:supp_impl} \\
\indent A.1. Details of Training Data \dotfill \pageref{subsec:supp_data_details} \\
\indent A.2. More Architecture Details \dotfill \pageref{subsec:more_arch_details} \\
\indent A.3. More Training Details \dotfill \pageref{subsec:more_training_details} \\
\indent A.4. More Inference Details \dotfill \pageref{subsec:more_inference_details} \\
\indent A.5. More Details on the Proposed Baseline \dotfill \pageref{subsec:more_baseline_details} \\
\vspace{-1em}

\noindent \textbf{B. More Experimental Results} \dotfill \pageref{sec:supp_exp} \\
\indent B.1. More Evaluation on 7-Scenes \dotfill \pageref{subsec:7scenes_eval} \\
\indent B.2. Depth Evaluation \dotfill \pageref{subsec:depth_eval} \\
\indent B.3. More Detailed Results on VBR \dotfill \pageref{subsec:more_vbr} \\
\indent B.4. More Qualitative Results on Trajectory \dotfill \pageref{fig:kitti_qual} \\
\indent B.5. More Qualitative Results on 3D Reconstruction \dotfill \pageref{fig:gallery} \\
\vspace{-1em}

\section{Additional Implementation Details}
\label{sec:supp_impl}
\subsection{Details of Training Data}
\label{subsec:supp_data_details}

To train our model to effectively handle long-context geometric reconstruction, we utilize a diverse mixture of 14 large-scale datasets, encompassing both real-world and synthetic scenes across indoor, outdoor, and autonomous driving environments. 

To standardize the inputs for our chunk-wise architecture, all datasets are processed into multi-view sequences consisting of $48$ views (or $128$ views for the training stage on H200 GPUs) with a uniform resolution of $504 \times 280$. The sampling strategy follows CUT3R~\cite{wang2025continuous}.
Furthermore, to ensure the quality of geometric supervision, we apply rigorous depth filtering across all datasets. Depending on the dataset characteristics (whether metric scale or not), we utilize either a maximum depth threshold (\eg, 80.0 meters for ARKitScenes and ScanNet) or a percentile-based clipping strategy (\eg, 90th or 98th percentile for DL3DV and TartanAir) to mask out noisy or invalid depth values.

As discussed in the main text, overcoming the ``data wall'' requires providing the model with sufficient long-horizon signals and diverse scene priors. To achieve this, we construct our training mixture by heavily weighting large-scale navigation datasets (\eg, TartanAirV2 and VKITTI2) to encourage long-range geometric reasoning. 
While DL3DV does not feature extremely long spatial trajectories like the navigation datasets, we also assign it a correspondingly high sampling weight due to its exceptional real-world scene diversity, which is critical for model generalization. 
Conversely, we down-weight smaller or object-centric datasets. 
Additionally, we note that at the time of training, only a subset of 5,000 sequences from the OmniWorld-Game dataset was publicly released, which comprises the portion we utilized in our mixture. The complete list of datasets and their relative sampling percentages is summarized in~\cref{tbl:data_mixture}.

\begin{table}[t]
\centering
\caption{\textbf{Training data mixture.} We list the datasets used in our final training configuration along with their relative sampling percentages. Navigation and large-scale scene datasets are heavily weighted to encourage long-range geometric reconstruction.}
\vspace{-0.5em}
\label{tbl:data_mixture}
\resizebox{1\columnwidth}{!}{
\begin{tabular}{lc}
\toprule
\textbf{Dataset} & \textbf{Percentage (\%)} \\
\midrule
DL3DV~\cite{ling2024dl3dv} & 17.89 \\
TartanAirV2~\cite{patel2025tartanground} & 17.89 \\
OmniWorld-Game (subset)~\cite{zhou2025omniworld} & 17.89 \\
ARKitScenes~\cite{baruch2021arkitscenes} & 10.44 \\
TartanAir~\cite{wang2020tartanair} & 8.94 \\
Waymo~\cite{schwall2020waymo} & 6.71 \\
ARKitScenes HighRes~\cite{baruch2021arkitscenes} & 4.18 \\
MegaDepth~\cite{li2018megadepth} & 4.18 \\
ScanNet~\cite{dai2017scannet} & 4.18 \\
ScanNet++~\cite{yeshwanth2023scannet++} & 3.13 \\
Virtual KITTI 2~\cite{cabon2020vkitti2} & 2.24 \\
HyperSim~\cite{roberts2021hypersim} & 2.08 \\
Spring~\cite{mehl2023spring} & 0.22 \\
UnReal4K~\cite{wang2023neural} & 0.04 \\
\bottomrule
\end{tabular}
}
\vspace{-1.5em}
\end{table}

\subsection{More Architecture Details}
\label{subsec:more_arch_details}

The base network architecture is similar to \pithree{}~\citep{wang2025pi}, comprising a DINO-based patchifier, 18 residual blocks, and dedicated prediction heads for the local pointmap, global camera pose, and confidence map.

We insert a TTT layer into each of the 18 residual blocks, while the SWA layers are sparsely integrated only at the 6th, 10th, 14th, and 18th blocks. These SWA layers are initialized from the corresponding global attention layers in \pithree{}. Furthermore, to provide temporal context within the SWA layers, we add three learnable positional embedding tokens to the frame representations. These tokens explicitly indicate whether a frame overlaps with the previous chunk, has no overlap, or overlaps with the subsequent chunk. 
The TTT layer features a head dimension of 512 and an intermediate layer with an expansion factor of 4. 
In terms of model size, the base network contains approximately 954M parameters, while the introduced TTT and SWA layers account for an additional 296M parameters.

\subsection{More Training Details}
\label{subsec:more_training_details}

\noindent \textbf{Optimization and Hyperparameters.}
We optimize our model using AdamW with momentum parameters $\beta=(0.9, 0.999)$. During training, we freeze the encoder and the prediction heads to retain their pre-trained feature representations. We apply decoupled learning rates: $5 \times 10^{-4}$ for the newly introduced TTT and SWA layers, and a smaller learning rate of $1 \times 10^{-5}$ for the unfrozen parameters of the base network. To prevent overfitting while maintaining training stability, we employ a selective weight decay strategy with a factor of $0.05$. Specifically, weight decay is exclusively applied to multi-dimensional parameter tensors (\ie $\ge 2$ dimensions, such as weight matrices in linear or convolutional layers). One-dimensional or scalar parameters, including biases and normalization scale/shift parameters (\eg, in LayerNorm), remain undecayed. This practice prevents the disruption of normalization statistics and preserves network expressivity. For the training objective, the loss weights are empirically set to $\lambda_{\text{r}}=0.1$ for rotation, $\lambda_{\text{t}}=10$ for translation, and $\lambda_{\text{global}}=1$ for the global pointmap.

\noindent \textbf{Training Schedule and Curriculum.}
We use a cosine learning rate decay schedule with a warmup of 1,000 steps. To progressively build the model's capacity for long-context reasoning, our curriculum training is split into two hardware-specific stages. In the first stage, we train the model for 25,000 steps on NVIDIA H100 GPUs using a fixed sequence length of 48 frames. Over this stage, we linearly decrease the chunk size (the number of frames per chunk) from 12 to 4, and the overlap size from 3 to 1, effectively increasing the number of recurrent steps. In the second stage, we leverage NVIDIA H200 GPUs to train for an additional 15,000 steps, extending the total sequence length to 128 frames. During this phase, we linearly decrease the chunk size from 12 to 8, and the overlap size from 3 to 2.

\noindent \textbf{Efficiency Implementations.}
Given the extreme memory demands of unrolling long sequences and maintaining recurrent states, we heavily rely on gradient checkpointing across the network blocks to reduce memory consumption. Furthermore, we implement our chunk-wise sliding window attention (SWA) using FlexAttention~\citep{dong2024flex} to optimize both memory footprint and compute efficiency during training.

\subsection{More Inference Details}
\label{subsec:more_inference_details}

\noindent \textbf{Inference Configurations.}
For the evaluation of short sequences and small-scale scenes, we employ a chunk size of 64 and an overlap size of 3 for both \ourmethod{} and \ourmethod{}*. For long sequences and large-scale environments, we maintain this exact configuration for \ourmethod{}*. However, because the base \ourmethod{} model is more susceptible to trajectory drift at larger scales, we reduce its chunk size to mitigate error accumulation. Specifically, for \ourmethod{}, we use a chunk size of 32 for the KITTI dataset and 48 for the VBR dataset.

\noindent \textbf{Latency and Efficiency.}
To optimize computational efficiency during inference, we implement a KV-cache mechanism for the block-wise SWA layers, avoiding redundant token computations. We evaluate the inference speed and peak memory consumption using a single NVIDIA A100 (40GB) GPU over a sequence of 500 frames. As detailed in~\cref{tbl:inference_efficiency}, the performance scales predictably with the chosen chunk size, demonstrating a clear trade-off between the temporal context window and computational cost. We believe that inference efficiency can be further improved by pruning additional TTT layers or by employing strided sampling within the SWA module for historical, non-overlapping frames. We leave these system-level optimizations for future work.

\begin{table}[t]
\centering
\small
\caption{\textbf{Inference efficiency.} Measured on a single NVIDIA A100 (40GB) GPU over a 500-frame sequence with varying chunk sizes. As the number of frames increases, the speed improves slightly, while memory consumption remains constant.}
\vspace{-0.5em}
\label{tbl:inference_efficiency}
\begin{tabular}{ccc}
\toprule
\textbf{Chunk Size} & \textbf{Speed (FPS)} & \textbf{Memory (GB)} \\
\midrule
64 & 9.3  & 27.2 \\
48 & 10.6 & 22.3 \\
32 & 12.1 & 18.1 \\
\bottomrule
\vspace{-2em}
\end{tabular}
\end{table}

\begin{table*}[t]
\centering
\caption{\textbf{Comparison of Absolute Trajectory Error (ATE$\downarrow$, m) on VBR.} The top and bottom blocks denote \textit{optimization-based} and \textit{feedforward} methods, respectively. \textbf{Bold} and \underline{underline} indicate the best and second-best performance among feedforward methods. 
}
\label{tbl:vbr}
\resizebox{\textwidth}{!}{%
\begin{tabular}{clccccccc|c}
\toprule
& \textbf{Methods} & colosseo\_0 & campus\_0 & campus\_1 & pincio\_0 & spagna\_0 & diag\_0 & ciampino\_1 & \textbf{Avg.} \\
& \scriptsize \textit{num. of frames, scale} & \scriptsize 8815, 1.45km & \scriptsize 12042, 2.73km & \scriptsize 11671, 2.95km & \scriptsize 11142, 1.27km & \scriptsize 14141, 1.56km & \scriptsize 10021, 1.02km & \scriptsize 18846, 5.20km & - \\
\midrule
\multirow{3}{*}{\rotatebox{90}{\textbf{Opt.}}}
& VGGT-SLAM~\citep{maggio2025vggt} & 102.92 & 110.77 & 89.69 & 72.98 & 62.67 & 35.62 & 144.02 & 88.38 \\
& VGGT-Long w/o LC~\citep{deng2025vggt} & 86.46 & 132.66 & 115.65 & 64.40 & 57.39 & 33.66 & 186.13 & 96.62 \\
& VGGT-Long~\citep{deng2025vggt} & 45.73 & 132.68 & 115.65 & 64.40 & 58.49 & 33.66 & 187.97 & 91.23 \\
\midrule
\multirow{6}{*}{\rotatebox{90}{\textbf{Feedforward}}}
& InfiniteVGGT~\citep{yuan2026infinitevggt} & 90.02 & 136.83 & 116.69 & 77.72 & 63.47 & 33.91 & 202.11 & 102.96 \\
& CUT3R~\citep{wang2025continuous} & 88.16 & 46.39 & 48.00 & 52.87 & 47.98 & 30.18 & 187.58 & 71.59 \\
& TTT3R~\citep{chen2025ttt3r} & 81.63 & 68.36 & 65.67 & 35.56 & 39.03 & \textbf{19.84} & 185.59 & 70.81 \\
\cmidrule{2-10}
& Pi3-Chunk (Proposed Baseline) & 81.63 & 86.54 & 71.49 & 49.85 & 54.27 & \underline{28.07} & 118.81 & 70.09 \\
& \cellcolor{gray!10}\textbf{\ourmethod{}} \textbf{(Ours)} & \cellcolor{gray!10}\textbf{25.60} & \cellcolor{gray!10}\textbf{19.53} & \cellcolor{gray!10}\textbf{28.90} & \cellcolor{gray!10}\underline{28.11} & \cellcolor{gray!10}\textbf{23.04} & \cellcolor{gray!10}35.01 & \cellcolor{gray!10}\underline{92.94} & \cellcolor{gray!10}\underline{36.16} \\
& \cellcolor{gray!10}\textbf{\ourmethod{}*} \textbf{(Ours)} & \cellcolor{gray!10}\underline{49.38} & \cellcolor{gray!10}\underline{22.44} & \cellcolor{gray!10}\underline{34.90} & \cellcolor{gray!10}\textbf{11.08} & \cellcolor{gray!10}\underline{27.05} & \cellcolor{gray!10}32.88 & \cellcolor{gray!10}\textbf{44.51} & \cellcolor{gray!10}\textbf{31.75} \\
\bottomrule
\end{tabular}%
}
\vspace{-0.25em}
\end{table*}

\begin{figure*}[t]
    \centering
    \begin{subfigure}[t]{0.85\textwidth}
        \includegraphics[width=\linewidth]{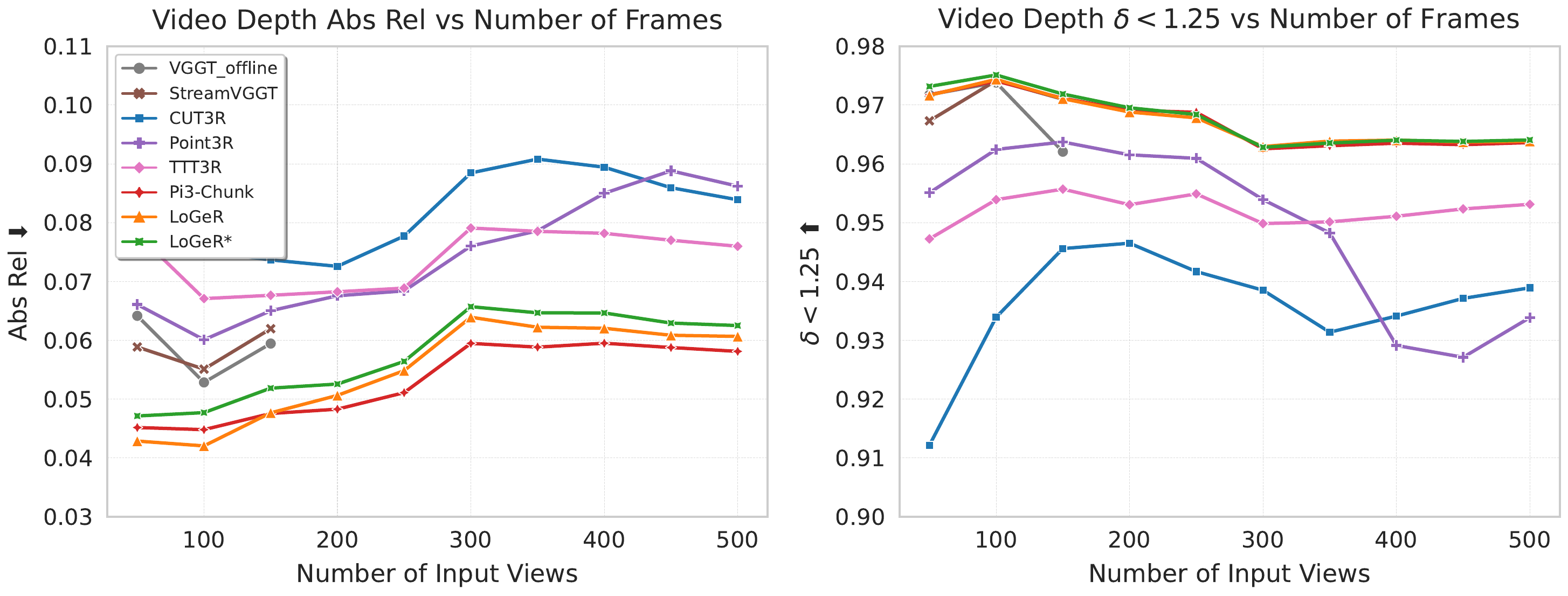}
    \end{subfigure}
    \vspace{-0.25em}
    \caption{\textbf{Quantitative results of video depth estimation on the Bonn dataset. }Both our proposed baseline (Pi3-Chunk) and \ourmethod{} significantly outperform prior works. Note that VGGT and StreamVGGT are omitted after certain frames due to out-of-memory (OOM).}
    \vspace{-0.5em}
    \label{fig:bonn}
\end{figure*}

\subsection{More Details on the Proposed Baseline}
\label{subsec:more_baseline_details}

Based on the philosophy of chunk-causal processing, we introduce \textit{Pi3-Chunk}, a simple baseline built on top of $\pi^3$~\citep{wang2025pi} for long sequence reconstruction. Specifically, we process the input sequence in a chunk-wise manner similar to \ourmethod{}, and then rely on a feedforward alignment step to stitch the predictions of different chunks together using their overlapping frames.

While the alignment in \ourmethod{}$^{\ast}$ strictly requires a rigid SE(3) transformation $\mathbf{A}_m$, \textit{Pi3-Chunk} necessitates a SIM(3) transformation. This is because the base $\pi^3$ model predicts geometries and translations that are only consistent up-to-scale within each individual chunk. Therefore, to align the current chunk $\mathcal{C}_m$ to the globally aligned coordinate system of the previous chunk $\mathcal{C}_{m-1}$, we must first recover and compensate for the relative scale shift. 

Let $\hat{\mathbf{x}}^{(m)}_{k,p}$ denote the raw predicted local point coordinates of the overlapping frame $k$ at pixel $p$ in the current chunk $\mathcal{C}_m$, and let $\tilde{\mathbf{x}}^{(m-1)}_{k,p}$ denote the corresponding \emph{scale-adjusted} local point coordinates from the previously aligned chunk $\mathcal{C}_{m-1}$. Because these pointmaps represent the same physical geometry in the camera coordinate system, their ratio directly reflects the scale ambiguity. We robustly estimate the relative scale factor $s_m$ by taking the median (or alternatively, a truncated mean, which yields similar empirical performance) of the point-wise coordinate norms over all valid pixels $p \in \Omega$:
\begin{align}
s_m = \mathop{\mathrm{Median}}_{p\in\Omega} \left( \frac{\| \tilde{\mathbf{x}}^{(m-1)}_{k,p} \|_2}{\| \hat{\mathbf{x}}^{(m)}_{k,p} \|_2} \right).
\label{eq:pi3_chunk_scale}
\end{align}

Once the scale factor $s_m$ is determined, we apply it to adjust the local pointmaps and the translation components of all raw poses in the current chunk. Let $\hat{\mathbf{T}}^{(m)}_{t} = [\hat{\mathbf{R}}^{(m)}_{t} \mid \hat{\mathbf{t}}^{(m)}_{t}]$ be the raw predicted pose. We obtain the scale-adjusted pose $\bar{\mathbf{T}}^{(m)}_{t}$ as:
\begin{align}
\bar{\mathbf{T}}^{(m)}_{t} = \big[ \hat{\mathbf{R}}^{(m)}_{t} \mid s_m \hat{\mathbf{t}}^{(m)}_{t} \big], \ \ \forall t \in \mathcal{C}_m.
\label{eq:pi3_chunk_scaled_pose}
\end{align}
Additionally, the local pointmaps for this chunk are recursively scaled as $\tilde{\mathbf{x}}^{(m)}_{t,p} = s_m \hat{\mathbf{x}}^{(m)}_{t,p}$ to serve as the reference for the next chunk.

Finally, similar to the strategy in \ourmethod{}$^{\ast}$, we compute the SE(3) alignment $\mathbf{A}_m$ using the scale-adjusted pose of the overlapping frame $k$:
\begin{align}
\mathbf{A}_m = \tilde{\mathbf{T}}^{(m-1)}_{k}\big(\bar{\mathbf{T}}^{(m)}_{k}\big)^{-1}.
\label{eq:pi3_chunk_alignment}
\end{align}
This rigid transformation is then applied to all scale-adjusted frames in the current chunk to obtain the final global camera poses:
\begin{align}
\tilde{\mathbf{T}}^{(m)}_{t} = \mathbf{A}_m\,\bar{\mathbf{T}}^{(m)}_{t},\ \ \forall t\in\mathcal{C}_m.
\label{eq:pi3_chunk_final}
\end{align}

\section{More Experimental Results}
\label{sec:supp_exp}

\subsection{More Evaluation on 7Scenes}
\label{subsec:7scenes_eval}

\begin{figure}[t] %
    \centering
    \includegraphics[width=1\linewidth]{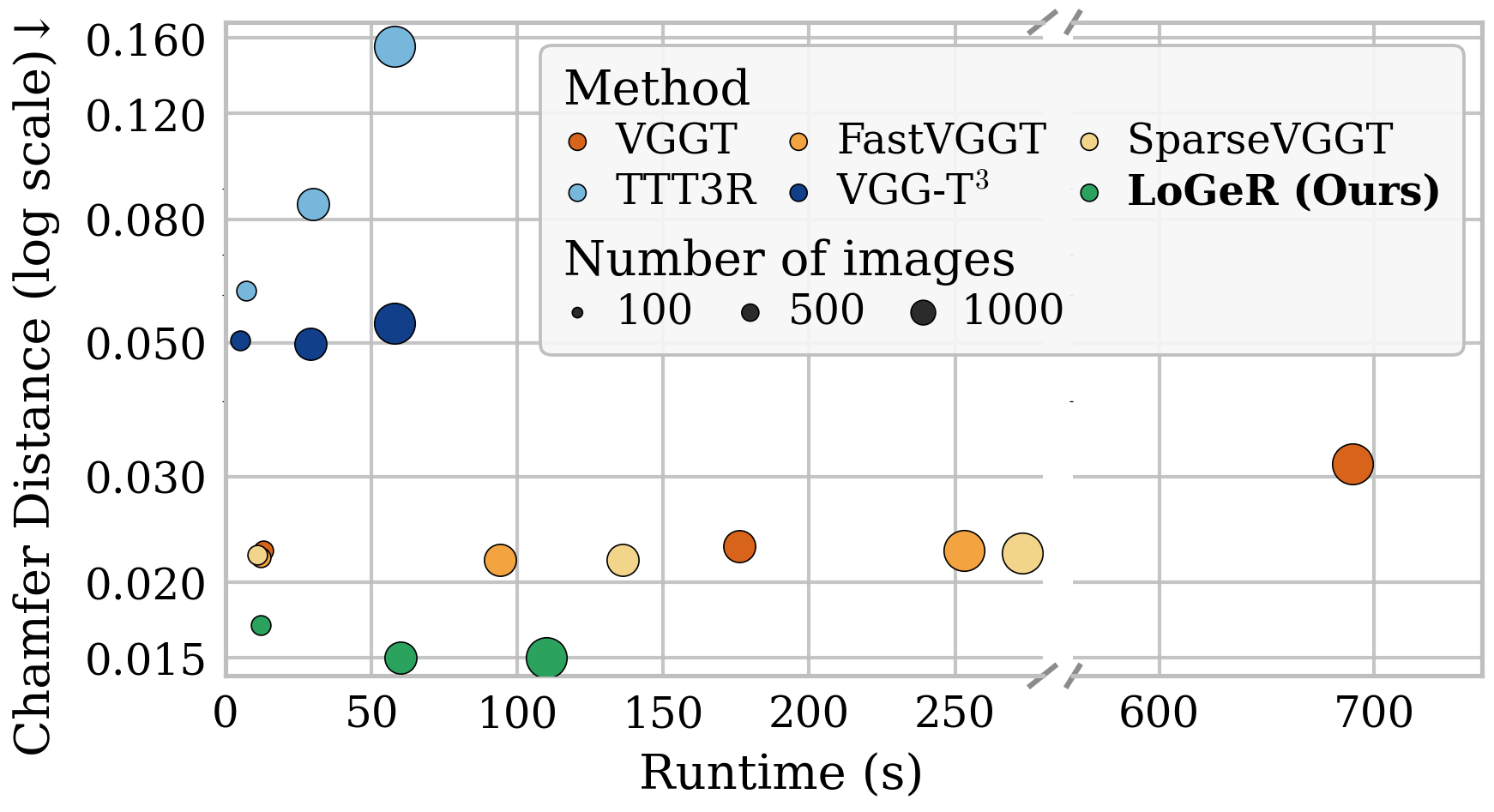} %
    \vspace{-1em}
    \caption{
        \textbf{3D reconstruction results on the 7-Scenes dataset}, uniformly sampling 100, 500, and 1k frames. For the evluation of 1k frames, our method achieves a \textbf{90.3\%} and \textbf{72.1\%} performance boost (error reduction) with comparable runtimes against concurrent works TTT3R~\cite{chen2025ttt3r} and VGG-T$^3$~\cite{elflein2026vgg}, respectively. \ourmethod{} also achieves \textbf{84.1\%} faster inference speed compared to VGGT with 31.0\% better performance. 
    }
    \vspace{-1em}
    \label{fig:uniform_7scenes}
\end{figure}

Following the evaluation protocol of VGG-T$^3$~\cite{elflein2026vgg}, we assess our method on the 7Scenes dataset by uniformly sampling frames across the sequences. As shown in~\cref{fig:uniform_7scenes}, \ourmethod{} significantly outperforms recent concurrent works across all sampled frame counts. 

 \renewcommand{\dbltopfraction}{0.95}
\renewcommand{\textfraction}{0.05}
\renewcommand{\dblfloatpagefraction}{0.85}

\begin{figure*}[t]
    \centering
        \includegraphics[width=0.91\linewidth]{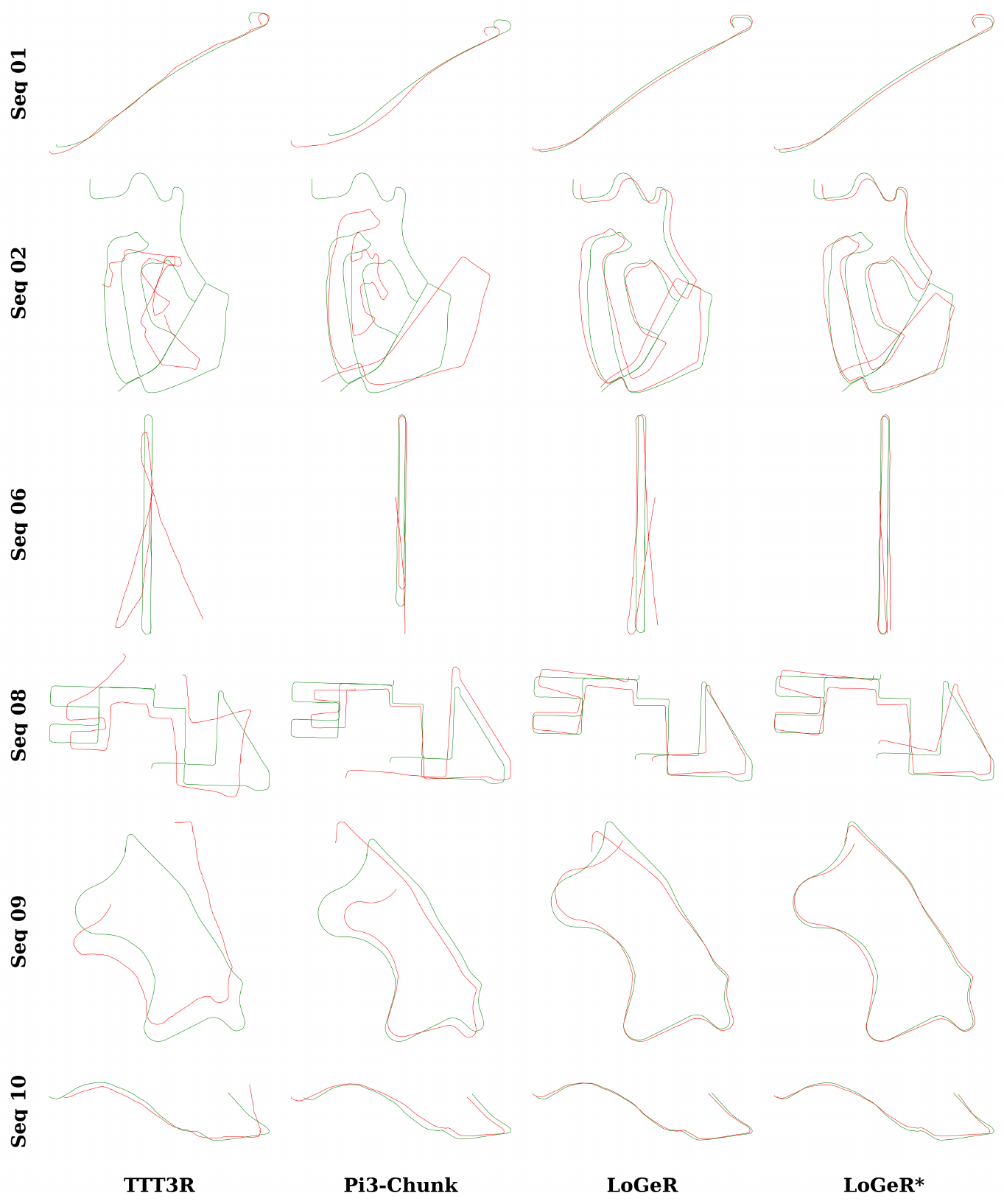}
    \caption{\textbf{Qualitative results on KITTI.} Ours shows more accurate long trajectory estimation.}
    \vspace{-1em}
    \label{fig:kitti_qual}
\end{figure*}

\subsection{Depth Evaluation}
\label{subsec:depth_eval}

We follow the evaluation protocol of TTT3R~\citep{chen2025ttt3r} to evaluate video depth on the Bonn dataset~\citep{palazzolo2019refusion} for sequences up to 500 frames. 

Conceptually, depth estimation is a more localized task compared to global camera pose estimation or 3D reconstruction, as it relies less on strict, long-term global consistency. 
Nevertheless, as shown in~\cref{fig:bonn}, both our proposed Pi3-Chunk baseline and \ourmethod{} significantly outperform prior methods. Specifically, \ourmethod{} achieves a $21.05\%$ error reduction (Abs Rel at 500 frames) compared to the previous best-performing method, TTT3R, demonstrating the efficacy of our methods.

\vspace{-0.2em}
\subsection{More Detailed Results on VBR}
\vspace{-0.2em}
\label{subsec:more_vbr}
We show in~\cref{tbl:vbr} the per-sequence result on VBR dataset. Our methods achieve the best on almost all sequences.

\subsection{More Qualitative Results on Trajectory}
\label{subsec:more_kitti}
\vspace{-0.2em}

We show in~\cref{fig:kitti_qual} the qualitative trajectory comparison on KITTI benchmark.

\vspace{-0.2em}
\subsection{More Qualitative Results on 3D Reconstruction}
\begin{figure*}[t]
    \centering
    \begin{subfigure}[t]{\textwidth}
        \includegraphics[width=\linewidth, trim=0 0 0 1em, clip]{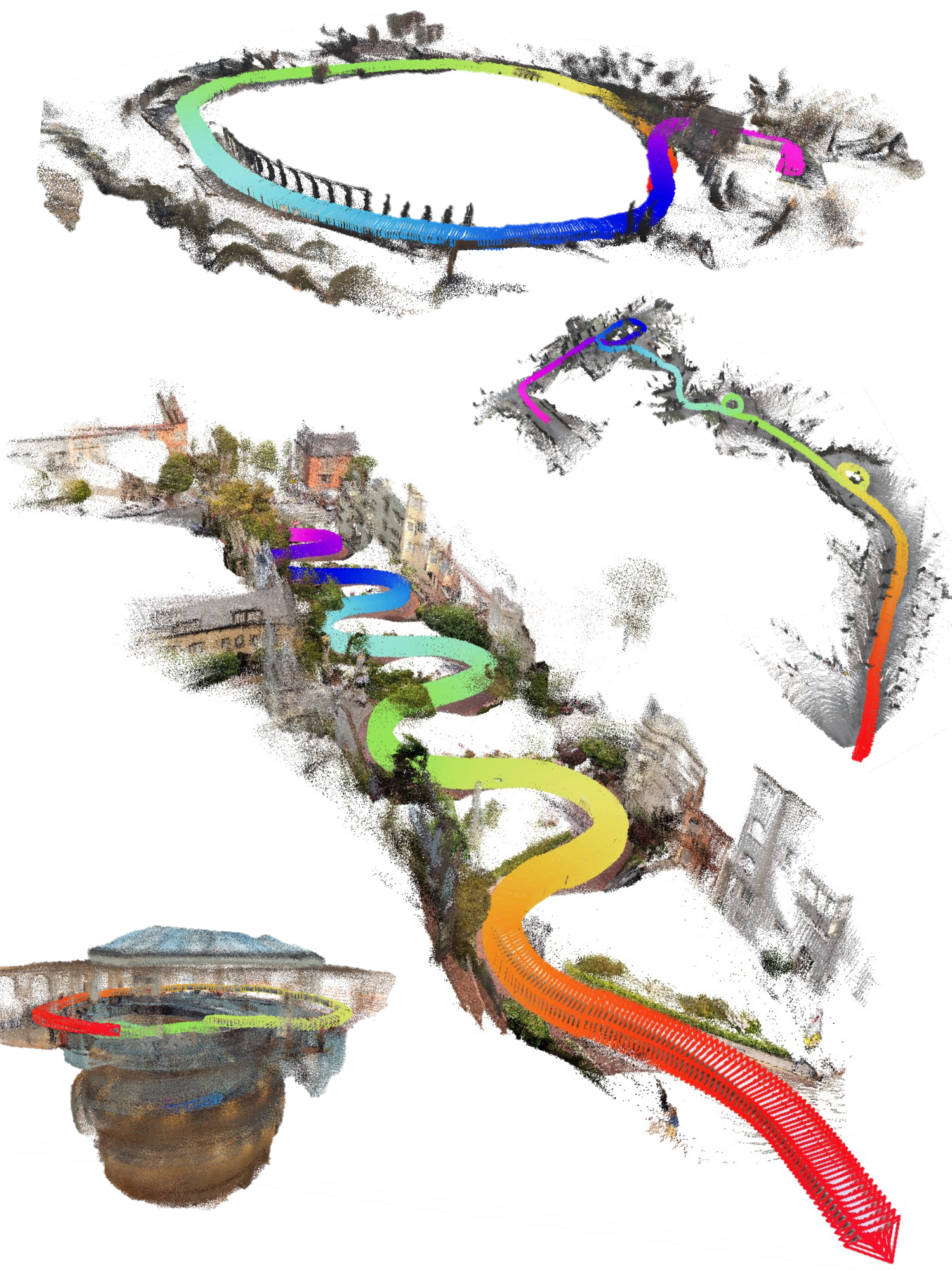}
    \end{subfigure}
    \caption{\textbf{Qualitative results on large-scale outdoor and indoor scenes}, showing minutes-long long-horizon reconstruction results.}
    \vspace{-1em}
    \label{fig:gallery}
\end{figure*}

\vspace{-0.2em}
\label{subsec:gallery}
\cref{fig:gallery} shows more results on large-scale indoor and outdoor scenes from in-the-wild videos and VBR.

\end{document}